# Use of Markov Chains to Design an Agent Bidding Strategy for Continuous Double Auctions

**Sunju Park**                                           SPARK@BUSINESS.RUTGERS.EDU
*Management Science and Information Systems Department*
*Rutgers Business School, Rutgers University*
*Newark, NJ 07102*

**Edmund H. Durfee**                                        DURFEE@UMICH.EDU
*Artificial Intelligence Laboratory, University of Michigan*
*Ann Arbor, MI 48109-2110*

**William P. Birmingham**                            WPBIRMINGHAM@GCC.EDU
*Math & Computer Science Department, Grove City College*
*Grove City, PA 16127*

## Abstract

As computational agents are developed for increasingly complicated e-commerce applications, the complexity of the decisions they face demands advances in artificial intelligence techniques. For example, an agent representing a seller in an auction should try to maximize the seller's profit by reasoning about a variety of possibly uncertain pieces of information, such as the maximum prices various buyers might be willing to pay, the possible prices being offered by competing sellers, the rules by which the auction operates, the dynamic arrival and matching of offers to buy and sell, and so on. A naïve application of multiagent reasoning techniques would require the seller's agent to explicitly model all of the other agents through an extended time horizon, rendering the problem intractable for many realistically-sized problems. We have instead devised a new strategy that an agent can use to determine its bid price based on a more tractable Markov chain model of the auction process. We have experimentally identified the conditions under which our new strategy works well, as well as how well it works in comparison to the optimal performance the agent could have achieved had it known the future. Our results show that our new strategy in general performs well, outperforming other tractable heuristic strategies in a majority of experiments, and is particularly effective in a "seller's market," where many buy offers are available.

## 1. Introduction

Electronic commerce at present is mostly driven by human interactions (humans decide what to buy, where to buy from, and how much they are willing to pay), but automated trading agents are increasingly being developed and deployed (Eriksson & Janson, 2002; Greenwald, 2003; Arunachalam et al., 2003). To a large extent, automated agents have been restricted to simpler types of e-commerce problems, where decisions about buying and selling, and about prices at which to buy and sell, are relatively formulaic (see for example, http://www.ebay.com). The construction of particular types of auctions and other mechanisms designed for e-commerce systems can also simplify agent decision problems by making it irrational, for example, for an agent to offer a bid price that is different from its true valuation for the object being exchanged (Vickrey, 1961).

However, there are many types of trading problems in e-commerce where decisions that an agent must make are much less formulaic. An example of such problems, which has recently been the focus of considerable research in research community (Wellman, 2003; Greenwald, 2003; He at al. 2002; Tesauro & Das, 2001), is the problem of bidding in a continuous double auction (CDA). The CDA is an auction variant (Friedman, 1993, page 8) in which buyers and sellers post their buy and sell offers asynchronously, and the auction will match a buy and a sell offer at any moment as long as the buy bid is higher than or equal to the sell offer. The CDA supports timely transactions and achieves high market efficiency, which is why most real-world markets for trading equities and commodities use the CDA. The dynamics of the CDA (such as entry and exit of agents' offers and continuous matching in auctions), however, adds additional complexity to the design of optimally profitable bidding agents. Although there has been research on developing bidding agents for CDAs (He et al., 2002; Tesauro & Das, 2001; He & Jennings, 2003; Vetsikas & Selman, 2003), there is no known dominant or equilibrium strategy in the CDA.

The goal of our work described in this paper has been to develop artificial intelligence techniques that enable the creation of high-performance bidding agents for CDAs. We have specifically investigated the following research questions:

- What is an agent's decision problem in the CDA?
- How should an agent's bidding strategy for the CDA be designed? That is, what kind of information should be used (and how should it be used) in an agent strategy for the CDA?
- How well does the developed agent bidding strategy perform in maximizing the agent's profit? In particular, when does it work and why does it work (and when and why does it not)?

In an auction, an agent needs to think about the participants it interacts with. A bidder in an eBay auction may not seem to explicitly consider each other participant as a separate entity, but he may want to estimate how popular the item is, what the final prices of similar items were in previous auctions, and how many other sellers of similar items might enter the market, all of which are associated with competition among bidders. By doing so, the bidder engages in thinking about the participants it interacts with at least in aggregate, if not as individuals.

A seller who tries to maximize profit by selling as many items at the highest prices possible, for example, needs to think not only about the auction protocol but also about potential buyers and other competing sellers. One reasonable strategy for a seller then would be to model what each individual agent thinks and will do, and use these models to figure out its best offer. In game theory, agents build a model of each other's possible moves and payoffs to find out their best moves (e.g., equilibrium strategies) (Kreps, 1990). Other researchers have designed an agent with a recursive model of what it thinks about what the opponent thinks, about what it thinks the opponent thinks about what, and so on (Gmytrasiewicz & Durfee, 2000; Vidal & Durfee, 2003).

Modeling each of the other agents, however, is often impossible or impractical in decision problems involving a large number of evolving participants. The CDA is an example of such complex problems (Friedman & Rust 1993; Durfee et al., 1997), as agents can join or leave the CDA at any time such that it is impossible for an agent to build an elaborate model of each of the other agents. Instead, therefore, we have sought an alternative representation and tractable algorithms for this type of multiagent decision problem, and based on our research we have hypothesized that *an agent using a stochastic model of the auction process can perform well in the CDA*. With the stochastic model, a seller can use some (subjective) probabilistic assessment about competitors' selling prices, rather than figuring them out by modeling other competing sellers' reasoning processes. As the CDA is dynamic, an agent that captures the overall dynamics should be able to perform well. Ignoring the other agents' (internal) strategic reasoning may sound like a risky approximation, but as the dynamics of the CDA are partly caused by strategic agents, the stochastic model reflects, although indirectly, the strategic behavior of participating agents.

To test our hypothesis, we have developed a Markov Chain stochastic modeling technique as part of an agent's bidding strategy, which we call the **p-strategy** (for the *p*ayment strategy). An agent with the p-strategy can take into account the dynamics and resulting uncertainties of the auction process using stochastic modeling of the auction process. Furthermore, we have empirically evaluated the performance of the p-strategy. Under various environmental parameters, we have compared the performance of the p-strategy agent to agents with plausible alternative heuristics. Furthermore, since it is impossible to exhaustively compare the p-strategy against all other conceivable strategies to assess how close it comes to the "best" strategy, we have compared the p-strategy against the *ex-post* optimal strategy to make this determination. The *ex-post* optimum is the maximum profit an agent may have achieved had it known exactly how the future would unfold. The *ex-post* optimum cannot be achieved in the real world since it is computed based on the *ex post* analysis of what happened in the auction, but serves as an upper-bound benchmark on performance.

The rest of this paper is organized as follows. The related work is reviewed in Section 2. In Section 3, we define the agent's decision problem in the CDA and describe the basic concepts behind the p-strategy. Section 4 then explains the details of the p-strategy. In section 5, we examine the performance of the p-strategy through systematic experimentation. Section 6 summarizes the lessons learned and identifies opportunities for further research.

## 2. Related Work

As observed by Milgrom and Weber (1982, page 1117), "... One obstacle to achieving a satisfactory theory of bidding is the tremendous complexity of some of the environments in which auctions are conducted..." Most of the classical auctions examined in auction theory are one-sided auctions with many simplifying assumptions, while continuous double auctions (CDAs) are less studied due to their complexity. McAfee and McMillan noted that "... Few results on the double auction exist, because of the difficulties of modeling strategic behavior on both sides of the market. ... The oral double auction, with the bids and offers openly called, is still more difficult to model because the process takes place over time and agents do not know what prices will be available if they wait instead of trading now..." (McAfee & McMillan, 1987, page 726).

In short, non-cooperative game theory has not been very successful in finding game-theoretic solutions in double auctions. When no equilibrium strategies are known (for example, in the CDA), then there are few guidelines agents can follow, and the question naturally arises of how an agent should behave in such an auction. Many researchers have focused on developing heuristic agent strategies in various auction settings (Lee, 2001; Park, Durfee, & Birmingham, 2000; Friedman & Rust, 1993; Andreoni & Miller, 1995; He et al., 2002; Tesauro & Das, 2001; He & Jennings, 2003; Vetsikas & Selman, 2003).

To address the question of what an agent should do, some researchers have developed a recursive model of other agents (i.e., what they think about what I think and so on) (Gmytrasiewicz & Durfee, 2000). While game-theoretic agents are ultra-smart and super-rational such that they can reason about the whole recursive hierarchy *ad infinitum* (Brandenburger & Dekel, 1993), the recursive modeling method (RMM) assumes that agents can build only a finite nesting of models. Using the RMM, we can characterize agents as 0, 1, 2, … level modelers: a 0-level agent who has no model of other agents; a 1-level agent who does model the other agents (as 0-level agents), a 2-level agent who models other agents as 1-level agents, and so on. Basically, RMM is a decision-theoretic approach to game theory and has been advocated as a practical solution concept for developing an agent strategy. Our p-strategy can be viewed as 1-level RMM, since it models the aggregate behavior of other agents, but not the other agents' internal reasoning.

Examples of heuristic agent design can also be found in learning agents. The recursive reasoning method combined with reinforcement learning has been adopted in designing agents for double auctions (Hu & Wellman, 1996; Wellman & Hu, 1998). Hu and Wellman have studied the self-fulfilling bias of learning agents in synchronized double auctions. They have observed that oversimplified learning is risky—it often leads to worse results than a (simpler) competitive strategy, and that the agents' initial beliefs (i.e., biases) play an important role in the final results of the auction. Their results and others (Hu & Wellman, 1996; Gmytrasiewicz & Durfee, 2000) show that thinking deeper is not always beneficial. These observations have motivated us to develop an agent strategy that does not explicitly model the other agents' internal reasoning processes.

Evolutionary algorithms (EA) have also been used to design agent strategies in continuous double auctions (Cliff, 1998; Oliver, 1998). Cliff has developed a ZIP (*zero-i*ntelligence *p*lus) agent, an extension of the zero-intelligence (ZI) agent (who bids its cost plus random markup) (Gode & Sunder, 1993). Although the amount of markup is still randomly decided from some given independent and identically distributed (IID) distribution, the ZIP agent is equipped with a heuristic method for deciding when to raise its markup (e.g., whenever the last bid was accepted). The results show that trading among ZIP agents converges rapidly to equilibrium prices. By using ZIP agents, Preist (1999) has developed an agent-based double auction mechanism that achieves efficient allocations by eliminating trades outside of equilibrium price. We use the simple ZI agents as one of the alternative heuristic strategies when evaluating the performance of our p-strategy.

Rust, Miller, and Palmer (Rust et al., 1993) have carried out a double auction tournament. A very simple "waiting in the background" trading strategy has emerged as the winner of the tournament. As the reader may have guessed, this type of agent, although successful in the tournament, is not adaptive to market activities and is vulnerable to copies of itself. Given this experience, we made sure to evaluate our agent strategy against itself as well.

Badea (2000) applied inductive logic programming to induce trading rules for a CDA. His agent identifies buy (or sell) opportunities from historical market data, which in turn becomes an input to a learning algorithm. Fuzzy logic has also been used to develop an agent strategy for the CDA (He et al., 2002). The agent employs heuristic fuzzy rules and fuzzy reasoning mechanisms in order to determine the best bid given the current market state. It can dynamically adjust its bidding behavior to respond to the changes in supply and demand. As shown in their experiments, adaptivity plays an important role in a good agent strategy for CDAs.

The IBM research group has developed two bidding strategies for CDAs, based on extensions of two published strategies (Tesauro & Das, 2001; Das et al., 2001). The first strategy is based on the zero-intelligence plus strategy (Cliff, 1998). The second strategy is based on the Gjerstad-Dickhaut algorithm (Gjerstad & Dickhaut, 1998), which uses recent market activity to compute the probability that a given bid or ask price will be accepted. Their experiments indicate that no single strategy outperforms the other, but their two strategies perform well against both human and agent opponents in CDAs.

Although which heuristic strategy will be most effective depends on the specific environment, the results reported in the literature repeatedly indicate that simple heuristics work as effectively as (and sometimes more effectively than) more sophisticated ones. When developing the p-strategy, we have abandoned more sophisticated recursive reasoning in favor of modeling only the auction process, hypothesizing that the dynamics of the CDA are more important than modeling other agents' internal reasoning processes. This design decision has also been influenced by the fact that acquiring information about individual agents in the CDA with entry-and-exit is difficult and that incomplete modeling may result in worse performance.

Methodologically, we employ Markov process theory (Bhat, 1972) to model the auction process. That is, the CDA with entry-and-exit is modeled as a Markov chain. Note that the information required in game theory and our stochastic-modeling method is different. In the former, all the information about available strategies and payoffs of all the agents involved should be known. The latter approach requires less information by ignoring the strategic reasoning processes of other agents. We hypothesize that this latter kind of strategy can be useful for systems where modeling dynamics is important (like the CDA), and we test our hypothesis by measuring the performance of our new Markov-model-based p-strategy across a battery of experiments.

## 3. The P-Strategy

We first define the agent's decision problem in a CDA. Then, we describe the p-strategy algorithm using a simple Markov chain as an example.

### 3.1. An Agent's Decision Problem in a CDA

The CDA allows buy bids and sell asks to be submitted and traded at any time in a trading period. A match is made when a buy bid is higher than a sell ask (a buyer is willing to pay at least as much as a seller is asking), and the transaction is completed at the clearing price. The clearing price can be either the average of the spread between the buy bid and the sell ask, or it could be simply one or the other of the bid price or the ask price. Among many variants of CDAs, we focus on the CDA with two standing queues (one for buyers's bids and the other for sellers' asks). An incoming buy or sell offer that does not get matched with standing offers will be stored in the appropriate queue and wait for a future match.

Formally, the agent's decision problem is to find an offer price $\rho$ that maximizes its expected utility (i.e., maximizes $\bar{u}(\rho)$). The agent's utility function is defined as

$$\bar{u}(\rho) = P_S(\rho) \times U(Payoff_S(\rho)) + P_F(\rho) \times U(Payoff_F(\rho)), \qquad (3.1)$$

where $P_S(\rho)$ and $P_F(\rho)$ denote the probabilities of success ($S$) and failure ($F$), respectively, and $Payoff_S(\rho)$ and $Payoff_F(\rho)$ denote the payoffs of $S$ and $F$, respectively, given offer price $\rho$. Success is when the agent gets a match for its offer. $P_S(\rho)$ and $P_F(\rho)$ add up to 1. Failure happens when the agent's offer does not ever get a match. For example, an agent may define an expiration time for its offer and consider it a failure when its offer expires without any match. In this paper, we assume that, rather than an explicit expiration time (deadline) for offers, a failure occurs when an agent's offer gets bumped out of a full queue of standing offers to make room for a more promising offer (as will be defined later in Section 4). The utility function, $U(\cdot)$, is monotonically increasing, since an agent prefers a higher payoff.

We use capital $S$ and $F$ as subscripts for *Success* and *Failure*, respectively (e.g., *Payoff$_S$* for the payoff of success), and *seller* and *buyer* as superscripts for the seller and the buyer, respectively (e.g., $\rho^{seller}$ for the seller's offer price). The subscripts and superscripts are often omitted when the context is clear. (The Appendix collects together our notational conventions in one place for easy reference.)

The seller's payoff for a successful transaction (*Payoff$_S^{seller}$*) is defined as the clearing price (*CP*) of the auction minus the cost (*C*) minus the delay overhead. We assume that the only significant overhead is in delay between the time of its offer and the time of the successful match (defined as $\Delta_S(\rho)$). We assume the time discount of a delay has cost of *c,* where *c* is a constant ($TD(\Delta_S(\rho)) = c \cdot \Delta_S(\rho),\ where\ c \geq 0$). That is,

$$Payoff_S^{seller}(\rho) = CP - C - TD(\Delta_S), \tag{3.2}$$

The clearing price is a function of the offer prices of the buyer and the seller being matched. The function returns a clearing price that is no higher than the buyer's offer price, and no lower than the seller's offer price. Our treatment in this paper does not require a particular function, which could give the buyer the entire surplus, the seller the entire surplus, or split it evenly between the two.

If the offer fails to clear at the auction, the seller may be worse off because of the delay. The payoff of failure (*Payoff$_F^{seller}$*) is minus the cost of the delay between when it made its offer and when failure occurred (defined as $TD(\Delta_F(\rho))$), assuming the value of failure is 0. That is,

$$Payoff_F^{seller}(\rho) = -TD(\Delta_F(\rho)). \tag{3.3}$$

Symmetrically, the buyer's payoffs of $S$ and $F$ can be defined. The buyer's payoff of a successful bid (*Payoff$_S^{buyer}$*) is defined as its valuation (*V*) minus the clearing price (*CP*) minus the time discount ($TD(\Delta_S(\rho))$). The payoff of failure (*Payoff$_F^{buyer}$*) is minus the time discount assuming the value of failure is 0. That is,

$$Payoff_S^{buyer}(\rho) = V - CP - TD(\Delta_S(\rho)), \tag{3.4}$$
$$Payoff_F^{buyer}(\rho) = -TD(\Delta_F(\rho)). \tag{3.5}$$

Note that we assume that buyers and sellers all have a delay cost of some constant *c*, but that each individual can have a different constant *c*.

The agent's decision problem formalized as in the above does not address the issue of the timing of a bid. That is, an agent is only concerned with *how much* to bid but not *when* to bid, so it submits a new bid right away. Ignoring when to bid is acceptable as long as we assume no cost for submitting a bid and that bids are not removed from the queue in a FIFO manner. With these assumptions, an agent should be able to solve its decision problem at every time step, and bid the best offer price (and override its previous bid) at every time step. Of course, if there is cost involved with bidding (whether it be the actual cost of submitting a bid or the cost of overriding the previous bid), the agent should consider when to bid as well as how much to bid. If the auction does not allow an agent make an inferior bid compared to its previous bid (that is, if there is a bid improvement constraint), for example, an agent needs to think about the tradeoffs of bidding right now against bidding later on. Timing then becomes an important decision factor. We do not consider the timing issue in this paper.

## 3.2. The P-strategy Algorithm

Figure 1 describes the p-strategy. The *p-strategy( )* function solves the optimization problem of maximizing the expected utility. For each offer price, an agent using the p-strategy builds a Markov Chain (MC) model, by computing (1) a set of initial auction states, (2) how the auction might proceed from the initial states, and (3) the transition probabilities among the MC states (step 1). The MC model is built using available information about the auction (such as standing offers and bids, history of the auction, etc.). Depending on the amount of information available, different MC models can be possible. Having the MC model ready, the agent computes the probabilities and the payoffs of $S$ and $F$ of its offer and therefore the utility value (step 2).

The p-strategy is a heuristic strategy. It models the dynamics of the CDA stochastically using a Markov Chain (MC), assuming that the auction behaves as a random process. This might sound like a risky engineering decision for a situation where agents may behave strategically. Later in Section 5, we test our agent strategy to confirm that, although the p-strategy might not always be "the" best strategy in all the possible environments, it works reasonably well in most cases.

The MC model captures the variables that influence the agent's utility value and the uncertainties associated with them. For instance, a seller is likely to raise its offer when there are many buyers or when it expects more buyers to come. The MC model takes those variables into account in the MC states and the transition probabilities. The MC model is described in detail in Section 4.

Finding the best offer price is essentially an optimization problem, where the implicit objective function is the agent's utility function. The objective function has no closed-form expression as the MC model is used to compute payoffs and probabilities. If we assume unimodality[1] of the objective function, we can use a univariate optimization technique, such as interval-reduction or interpolation methods, to find a local maximum, which by theorem is the global optimum (Gill et al., 1981). Proving the unimodality of a function, however, is hard. Although the plot of the (implicit) utility function appears unimodal for the auctions we have tried, we are not able to prove it. We may assume unimodality in the future, given that the curves have always been unimodal in hundreds of tests we have done by enumerating prices. At present, however, we employ total enumeration; we compute the utility values for each possible offer price and choose the one with the highest expected utility. In practical implementations, most optimization algorithms are terminated with a reasonable degree of approximation, and in our auction domain, the degree of optimality we can achieve is inherently limited by the denomination of the prices and payoffs (penny, for example). Therefore, even with the discretization of prices and payoffs, we can still achieve a reasonable degree of optimality.

---

**Function** p-strategy() **returns** an offer price
    $\rho$: a range of possible offer prices
    *best_offer*: best offer price
    *max_util*: maximum utility value
  *best_offer* = 0; *max_util* = 0;                /* initializes the variables */
  **for** each $\rho$
    build a Markov chain and compute the transition probabilities;    (step 1)
    *util* = compute the utility value;                       (step 2)
    /* if the new *util* is higher than *max_util*, update the best offer */
    **if** (*util* > *max_util*)
      *best_offer* = $\rho$; *max_util* = *util*;
    **end**
  **end**
  **return** *best_offer*

**Figure 1: The p-strategy algorithm.**

---

[1] Unimodality is often used as a condition that ensures the existence of a proper (i.e., single, relative) maximum in a given interval when only function values are available. *f(x)* is unimodal in [*a*, *b*] if there exists a unique value *x\** ∈ [*a*, b] such that, given any $x_1$, $x_2$ ∈ [a, b] for which $x_1 < x_2$: if $x_2 < x^*$ then *f(x₁) < f(x₂)*; if $x_1 > x^*$ then *f(x₁) > f(x₂)* .

The complexity of the p-strategy algorithm is $O(\rho \cdot n^3)$, where $\rho$ is the number of possible prices and $n$ is the number of MC states. Step (1) of building a MC model and computing transition probabilities (among $n$ MC states) takes $O(n^2)$ time. Step (2) of computing the utility value takes $O(n^3)$ time due to matrix multiplication and inversion. The time complexity of the fastest matrix multiplication algorithm is $O(n^{2.831})$, but our implementation uses the standard matrix multiplication method, whose time complexity is $O(n^3)$ (Press et al., 1988). Note that even if we apply a polynomial interpolation method (instead of total enumeration) while assuming unimodality, the worst-case time complexity would still be $O(\rho \cdot n^3)$, although the average-case would be constant-rate faster.

## 4. The Markov Chain Model for the CDA

The MC model is valuable to bidding agents. Intuitively, agents with a complete model of other agents will perform better, but without repeated encounters a complete model is unattainable. In a CDA with entry-and-exit, an agent in its lifetime meets many agents, and as a result its model of other agents tends to be incomplete. In such a case, modeling the auction process using readily available information (such as bid history) is more suitable than modeling the interior reasoning of each agent participating in the auction. Even when information about each individual agent is available, such a model is usually too complex to be practical. For instance, the complexity of solving the RMM recursive model that has been developed down to level $l$ is $a^n \times \{(m^{l+1} - 1)/(m - 1)\}$, where $a$ = number of alternative offer prices for an agent, $n$ = number of interacting agents, $m$ = number of alternative models of other agents considered, $l$ = level to which recursive model is developed (Gmytrasiewicz & Durfee, 2000). In addition, we need to consider the complexity of building the model in the first place, which is larger than the complexity of solving it (Gmytrasiewicz & Durfee, 2000).

This section describes how to build the MC model and compute its transition probabilities for the CDA (step 1 in Figure 1), and how to compute the expected utility value from the MC model (step 2 in Figure 1). In the CDA where the clearing price is determined at the seller's offer price, sellers have somewhat stronger incentives to bid above cost, as they always set the market price and this affects their tradeoffs between the probability of trading and the profit earned (Kagel & Vogt, 1993, page 288). Therefore, we take the perspective of the p-strategy *seller* when explaining the MC model. Of course, buyers have incentives, although somewhat weaker, to behave strategically as well (i.e., to bid below valuation), and the p-strategy buyer's MC model can be constructed in a similar fashion.

In Section 4.1, we describe the variables captured in the MC model. Using this information, a p-strategy agent can build the MC model for each offer price. It determines the set of initial states (described in Section 4.2), models how the auction will proceed from the set of initial states (described in Section 4.3), and computes the transition probabilities between the MC states (described in Section 4.4). Once the MC model is built, the p-strategy finds the expected utility (described in Section 4.5). Note that although we describe the transitions from a start state to a set of initial states (in Section 4.3) and from intermediate MC states to other intermediate MC states (in Section 4.4) separately, both transitions can be computed in the same way as shown in Section 4.4.

## 4.1. Variables Captured in the MC Model

When determining the best offer price, a p-strategy seller needs to capture in its MC model the variables that influence the expected utility value. We divide those variables into three groups, as shown in Table 1. Depending on the available information, one could use a different set of information for the agent strategy. The variables in Table 1 are the information readily available in most CDAs, and we chose to capture all of them in the MC model.

The variables in the first group capture information about the current status of the auction. Those variables are used to determine the set of possible initial states. The amount of information available to the p-strategy seller can vary, but we identify three pieces of information: the number of standing offers in the auction, the probabilistic distribution[2] of standing offer prices, and the clearing-price quote. In Section 4.2, we show an example of how MC models would differ if different amounts of information were used.

The variables in the second group capture the history of the auction, which is used to model the future auction process. We extrapolate from the historic information to make predictions about the near future of the auction. Such expectations can be justified on two grounds. First, although each individual agent's behavior may be different and the demography of agent population may evolve in the long run, the aggregate behavior of agents, such as arrival rates and probability distributions of offer prices, can capture the behavior of the whole agent population, and this aggregate behavior will not change drastically in the short run modeled in the MC model. Secondly, when the aggregate behavior eventually changes, the p-strategy agent can update the historic information and revise its MC model accordingly.

Therefore, note that our expectations are based on a couple of assumptions. First, we assume that the agents act independently. For example, the buyers do not all get together and decide to hold back their offers until the sellers lower prices, at which point the buyers flood the market. Buyers acting independently give some stability. Second, the duration of a particular "short-run" episode for which the MC model is being generated is short compared to the rate of agent turnover. This enables the p-strategy to keep up with the gradual changes in the agent population. The p-strategy may not work as well when the previous assumptions are not met and therefore the auction becomes more volatile.

Finally, in addition to the information about the auction, the p-strategy seller needs information about itself—its offer price and its cost.

| Variables used by the p-strategy seller | | |
|---|---|---|
| Information about the auction | Current auction status | Number of standing buy offers |
| | | Number of standing sell offers |
| | | Probability distribution of standing offer prices |
| | | Next clearing-price quote |
| | Auction history | Arrival rates of buy offers |
| | | Arrival rates of sell offers |
| | | Probability distribution of buy offer prices |
| | | Probability distribution of sell offer prices |
| Information about self | | Its offer price |
| | | Its cost |

**Table 1: The variables used by the p-strategy seller to build the MC model and to compute its transition probabilities.**

---

[2] How to compute the probabilistic distribution of standing offer prices is explained in detail in Section 4.4.

## 4.2. Determining the Set of Initial States

Using the information about the current status of the auction, the p-seller can determine the set of possible initial states when it offers $\rho$. Depending on the amount of information available to the p-seller, the set of possible initial states may differ. In the following, we examine three cases with different amounts of information.

Let us first examine how to compute the set of initial states when the p-seller knows the number of standing offers and their offer-price distribution. To represent standing offers in the CDA, we use *(bb…bsss…s)* notation. For example, the *(bbs)* state represents two standing buy-bids and one standing sell-offer. When the p-seller submits its offer (represented as $s^p$) at the *(bbs)* state, there are four possible next auction states as follows:

- *($s^p bbs$)* (when the p-seller's offer is less than the lowest of the standing buy offers),

- *($bs^p bs$)* (when its offer is in between the two standing buy offers),

- *($bbs^p s$)* (when its offer is higher than the buy offers but less than the sell offer),

- *($bbss^p$)* (when its offer is higher than all the standing offers).

In the first two cases, *($s^p bbs$)* and *($bs^p bs$)*, the initial state will be the *Success* state as $s^p$ and the first *b* to its right get matched (see Figure 2-(a)).

If the next clearing-price quote is known in addition to the number of standing offers and the distributions, the computation of initial states becomes simpler. The clearing-price quote indicates the highest standing buy-bid (denoted as *b'*). If the offer price of the p-seller is lower than the quoted clearing price (i.e., $\$(s^p) \leq \$(b')$, where $\rho = \$(s^p)$)[3], the initial state will be the *Success* state. Otherwise, the initial state will be either *($bb's^p s$)* or *($bb'ss^p$)* (see Figure 2-(b)).

When the p-seller knows the exact prices of standing offers (instead of their distributions), the initial state will be a single state depending on its offer price. For example, when two buy offers and one sell offer are standing at the auction, and their offer prices are known, the p-seller can determine its initial state with the offer price $\rho$ with a probability of 1 (see Figure 2-(c)). The p-seller may offer a price that guarantees a successful match or decide to bid higher in the hope that it may achieve a higher profit by waiting for a higher buy offer to arrive, at the risk of a higher probability of failure because a higher bid is more likely to be bumped to make room for more reasonable bids.

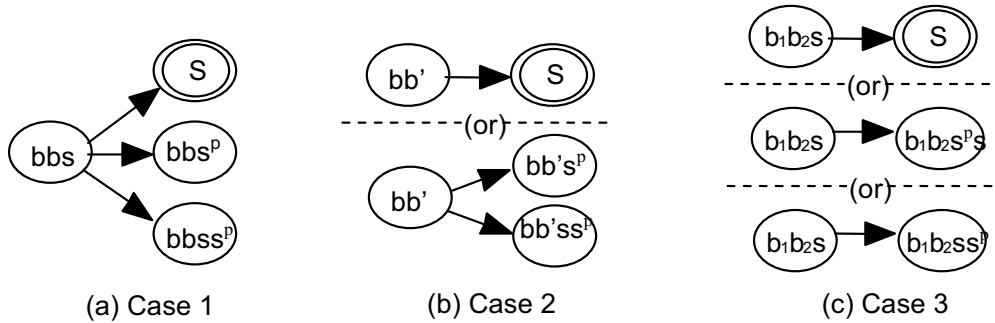

(a) Case 1          (b) Case 2          (c) Case 3

**Figure 2: The set of initial states from the *(bbs)* state depending on different information available.**

---

[3] We use *$(.)* notation to represent the value of the offer explicitly.

Note that when more information about the auction is used, the p-seller can predict the initial states more accurately, but the number of MC states becomes larger because it needs to keep track of more distinctive states. If it captures both the clearing-price quote (*b'*) and a new incoming buy offer (*b*) in the MC model, for example, the p-seller needs to distinguish between *(b'b)* and *(bb')* states. Which information to capture in the MC model is a design choice. Including the clearing-price when determining the initial set of states is an obvious choice, as the clearing price makes it possible to determine the probability and the payoff of success of the agent's offer right away. The value of including individual standing offer prices, however, is not that obvious. Given that the MC model of the rest of the auction process is built based on the probabilistic predictions on incoming offers (i.e., it uses distributions anyway), the exact values of standing offers become less useful. Moreover, modeling the individual standing offers would make the size of the MC model bigger. Thus, to manage the size of MC models, we have decided to use the clearing-price quote and the distributions of standing offers (and not individual standing offers). This corresponds to Case 2 in Figure 2-(b).

### 4.3. Modeling the Auction Process

From the set of initial states, the p-seller models how the auction will proceed. It is reasonable to assume that at most a single offer arrives to the auction at a time, because the CDA queues up offers that arrive simultaneously and tries to match each new offer (that is, clear offers) one at a time. We use the term clearing interval to denote the interval between successive attempts by the auction to match buy and sell offers. Assuming that offers arrive at most one at a time, the auction can go to any of the following states from the *(bbss$^p$)* state.

- *(bbss$^p$):* No offer arrives during the clearing interval.

- *(bbs$^p$):* A buy-bid arrives and is matched with the lowest sell-offer.

- *(bbbss$^p$):* A new buy-bid becomes a standing offer because of no match.

- *(bss$^p$):* A sell-offer arrives, and it is matched with the highest buy-bid.

- *(bbss$^p$s):* Because of no match, a new sell-offer becomes the highest standing offer.

- *(bbsss$^p$):* A new sell-offer becomes a standing offer, but the p-strategy agent's offer is still the highest.

All the remaining state transitions can be built in a similar way, and the resulting MC model is shown in Figure 3 From the current state of the auction (*(bbs* in this example), the process transitions to a set of possible initial states (marked in gray), and then to other states, and so on, until it goes to either the *S* or *F* state. The CDA modeled in Figure 3 limits the number of standing buy and sell offers not to exceed five each. As a result, a p-strategy seller will get bumped out (to the Failure state) when its offer is standing in a full standing-offer queue and a lower sell-offer arrives (see the *(sssss$^p$)* state, for example). When building the MC model for the CDA, we have made the design choice of using the clearing-price quote when determining the initial states, but not using that extra information when modeling the auction process afterwards. As the MC model of the rest of the auction process is based on the probabilistic predictions on incoming offers (i.e., it uses distributions), the exact value of the clearing price in the past would not be very useful. This compromise can be justified in terms of balancing the tradeoffs between more accurate initial-state computation and the explosion of the size of the MC model.

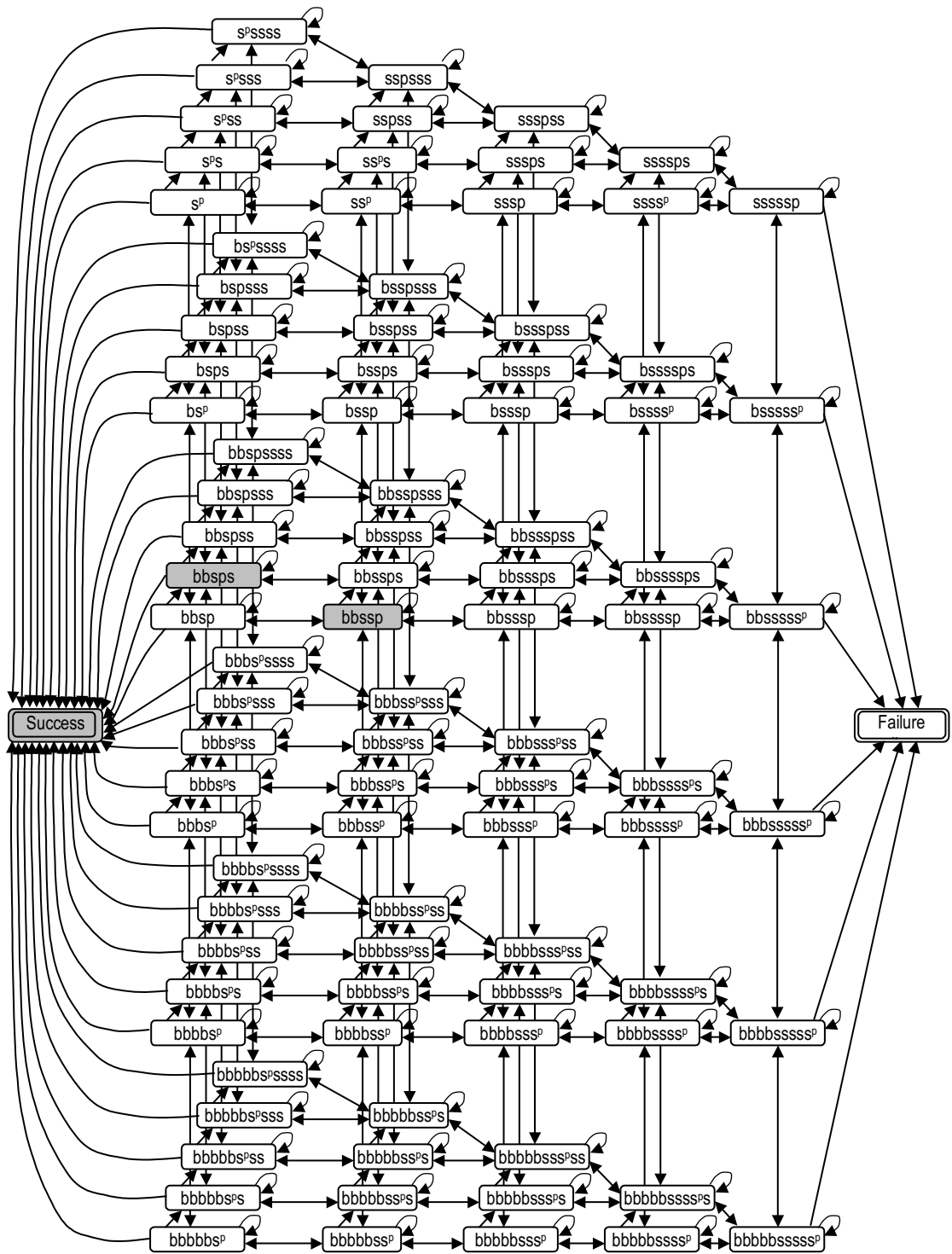

**Figure 3: The MC model of the CDA with starting state *(bbs)*.**

### 4.4. Computation of the Transition Probabilities

To complete the MC model, the p-seller needs to compute the transition probabilities between the MC states. Note that although we describe building the MC model and computing its transition probabilities separately, they are in fact a one-step process: the p-strategy agent computes the transition probabilities while building the MC model. As we assume the current clearing price, the highest standing buy-bid, is available, we treat it as constant when computing the transition probabilities.

We use the transition from the starting *(bbs)* state to *(bbss^p)* state and the transition from the *(bbss^p)* state to *(bbs^p)* state as our two examples to illustrate the process. Although we present the transition probability computation in two cases—the transition from a starting state (when it receives information about the auction) to a set of initial states and the transition between the MC states, both are in essence the same. The difference is the kind of information being used. When computing the transition probabilities from the starting state to the set of initial states, the p-seller uses information about the current status of the auction, such as the distributions of buy and sell offers and the number of standing offers. The transition probabilities between the MC states, on the other hand, are computed using the historic information about the auction dynamics (such as the arrival rates of buyers and sellers and the probabilistic distributions of their offer prices), which the p-strategy agent keeps track of.

Let transition probability $P_{ij}$ represent the conditional probability that the process in state $i$ makes a transition into state $j$. Let $X_n = i$ represent that the process is in state $i$ at time $n$. The transition probability from the starting *(bbs)* state to the *(bbss^p)* state, for example, then is:

$$P\{X_{n+1} = (bbss^p) \mid X_n = (bbs)\}, \quad \text{for all } n \geq 0. \tag{4.1}$$

Let $\$(s^p)$ denote the p-seller's offer price ($\$(s^p) = \rho$). The transition in (4.1) happens when the p-seller's offer ($s^p$) is higher than the standing sell-offer ($\$(s^p) > \$(s)$). That is,

$$P\{X_{n+1} = \$(b) \leq \$(b) \leq \$(s) \leq \$(s^p) \mid X_n = \$(b) \leq \$(b) \leq \$(s)\}. \tag{4.2}$$

Using Bayes' rule, (4.2) is re-written as:

$$P(X_{n+1} = \$(b) \leq \$(b) \leq \$(s) \leq \$(s^p) \ \& \ X_n = \$(b) \leq \$(b) \leq \$(s)) / X_n = P(\$(b) \leq \$(b) \leq \$(s)). \tag{4.3}$$

The second term in the numerator can be omitted, since it is always true when the first term is satisfied. Therefore, the transition probability is

$$P(X_{n+1} = \$(b) \leq \$(b) \leq \$(s) \leq \$(s^p)) / P(X_n = \$(b) \leq \$(b) \leq \$(s)). \tag{4.4}$$

That is, the transition probability from the current state of the auction (the *(bbs)* state in this example) to an initial state (the *(bbss^p)* state) for the p-strategy agent can be computed by finding two probabilities.

Let $f_b(b)$ [$f_s(s)$] be the probability density function (PDF) of a buy-bid [sell-ask]. If we assume the variables—buy-bids and sell-offers—are independent, the joint density function is the product of individual density functions. That is,

$$f(b,s) = f_b(b) \cdot f_s(s), \quad \text{for all } b \text{ and } s. \tag{4.5}$$

Equation (4.4) is then

$$\frac{\int_{s_1}^{s_2} \int_{b_1}^{s} \int_{\rho}^{B^2} f(S, B^2, B^1) dB^1 dB^2 \, dS}{\int_{s_1}^{s_2} \int_{b_1}^{s} \int_{b_1}^{B^2} f(S, B^2, B^1) dB^1 dB^2 \, dS} = \frac{\int_{s_1}^{s_2} \int_{b_1}^{s} \int_{\rho}^{B^2} f_s(S) f_b(B^2) f_b(B^1) dB^1 dB^2 \, dS}{\int_{s_1}^{s_2} \int_{b_1}^{s} \int_{b_1}^{B^2} f_s(S) f_b(B^2) f_b(B^1) dB^1 dB^2 \, dS}, \tag{4.6}$$

where $b_1$ is the lower bound of buy offers and $s_1$ and $s_2$ are the lower and the upper bounds of sell offers, respectively, and $\rho$ is the p-seller's offer price. Although this looks complex, the transition probability computation is a simple, repeated integration.

The transition probability from the *(bbss$^p$)* state to the *(bbs$^p$)* state happens when a new buy-bid arrives at the auction and matches the lowest sell-offer. That is,

$$P(bbs^p \mid bbss^p) = \frac{P\{(newbuyoffer(b') \, arrives) \, \& \, (b \le b \le s \le s')\}}{P(b \le b \le s \le s^p)} \qquad (4.7)$$

Let $\rho$ be the p-seller's offer price, $f(b)$ $[f(s)]$ be the probability density function (PDF) of the buyer-bid [sell-offer] distributions, and $p\_b$ $(p\_s)$ be the arrival rate of buy-bids [sell-offers].

The first term in the numerator, *P(newbuyoffer(b')arrives)*, is *p\_b*. The second term in the numerator and the denominator can be computed through repeated integration as before. That is,

$$P(b \le b \le s \le s^p) = \int_{-\infty}^{p} \int_{-\infty}^{p} \int_{-\infty}^{\phi} f_{B_1, B_2, S_1}(\xi_1, \xi_2, \xi_3) d\xi_1 d\xi_2 d\xi_3. \qquad (4.8)$$

## 4.5. Computing the Expected Utility Value

This section explains in detail how to compute the utility value for a given Markov chain using a simple Markov chain as an example. Representing an absorbing Markov chain using a canonical representation, computing the fundamental matrix, and dividing an absorbing MC into multiple Markov chains are standard techniques. In this section, we show how to apply these known techniques to develop a new way of computing the probabilities of success and failure and the payoffs of success and failure (therefore computing the expected utility of an offer price).

The example Markov chain in Figure 4 has two absorbing states (states 1 and 2) and two transitional states (states 3 and 4). State 1 is the Failure state, state 2 is the Success state, and state 3 is the initial state. For the MC model with multiple initial states, we can add a dummy initial state that transitions to the original multiple initial states and use the same following computation.

In Figure 4, the Markov model says that, if it offers a bid *x*, the agent can get a match (it goes to the *Success* state) with a probability of 0.8. Or, it does not get a match and waits for another incoming offer (it goes to State 4) with a probability of 0.2. From State 4, it gets a match with a probability of 0.1 (it goes to the *S* state), gets no match and waits for incoming offers with a probability of 0.5 (it stays at state 4), or it goes to the *Failure* state with a probability of 0.4.

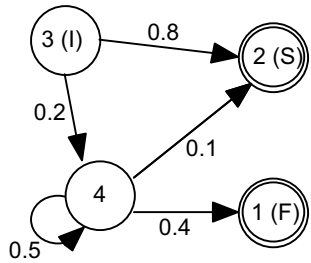

**Figure 4: A simple Markov chain (used as an example).**

Following standard definitions (Bhat 1972), a transition probability matrix, **P**, denotes a matrix of transition probabilities from state *i* to *j* ($P_{ij}$). Figure 5-(a) shows a canonical representation of an *(r+s)*-state MC consisting of *s* transient states and *r* absorbing states: **I** is an *(r×r)* identity matrix (each absorbing state transitions to itself); **O** consists entirely of 0's (by definition, an absorbing state never transitions to a transient state); **Q** is an *(s×s)* submatrix that captures the transitions only among the transient states; and $\boldsymbol{R}$ is an *(s×r)* matrix that represents the transitions from transient to absorbing states.

Figure 5–(b) shows the representation of the transition probability matrix of the Markov chain in Figure 4.

$$
\mathbf{P} = \begin{array}{c} \\ r \\ \\ s \end{array} \left[ \begin{array}{c|c} \overset{r}{\mathbf{I}} & \overset{s}{\mathbf{O}} \\ \hline \mathbf{R} & \mathbf{Q} \end{array} \right]
\qquad
\begin{array}{c|cccc}
 & 1 & 2 & 3 & 4 \\
\hline
1 & 1 & 0 & 0 & 0 \\
2 & 0 & 1 & 0 & 0 \\
\hline
3 & 0 & 0.8 & 0 & 0.2 \\
4 & 0.4 & 0.1 & 0 & 0.5
\end{array}
$$

(a) Canonical representation         (b) An example

**Figure 5: The transition probability matrix.**

In the real MC model, the transition probabilities are functions of the offer price. That is, the MC model is a *controlled* Markov chain[4] where the control input is the offer price. For simplicity, however, we use numeric values for transition probabilities in our example.

To find the utility value of its offer price ($u(\rho)$), an agent needs to compute the probabilities and the payoffs of $S$ and $F$. To compute the probabilities of $S$ and $F$, we use the fundamental matrix (defined shortly) of the original transition-probability matrix. To compute the payoffs of $S$ and $F$, on the other hand, we need to build two separate Markov chains from the original one, each of which has a single absorbing state (either Success or Failure), and re-compute the transition probabilities for each new MC from the original matrix. Then, from the new MC with the $S$ [$F$] state, the p-strategy agent computes the payoff of $S$ [$F$].

According to standard Markov chain theory (Bhat 1972), the fundamental matrix is defined as follows. Let $N_{ij}$ be the total number of times that the process visits transient state $j$ from state $i$. Let $N_{ij}^{k}$ be 1 if the process is in state $j$ after $k$ steps from state $i$, and 0 otherwise. Then, the average number of visits to state $j$ from state $i$ before entering any absorbing state is $E[N_{ij}]$.

$$E[N_{ij}] = E[\sum_{k=0}^{\infty} N_{ij}^{k}]$$

$$= \sum_{k=0}^{\infty} E[N_{ij}^{k}]$$

$$= \sum_{k=0}^{\infty} \{(1 - P_{ij}^{k}) \cdot 0 + P_{ij}^{k} \cdot 1\} \text{ (where } P_{ij}^{k} \text{ is the k-step transition probability)}$$

$$= \sum_{k=0}^{\infty} P_{ij}^{k}$$

$$= \sum_{k=0}^{\infty} \boldsymbol{Q}^{k} \text{ (since } i, j \text{ are transient states)} \tag{4.9}$$

From the transition probability matrix of any absorbing MC, the inverse of $\boldsymbol{(I - Q)}$ *always* exists (Bhat 1972), and

$$(I - Q)^{-1} = I + Q + Q^2 + \ldots = \sum_{k=0}^{\infty} Q^k \ . \tag{4.10}$$

---

[4] The transition probability from state $i$ to state $j$ of a controlled Markov chain with control input $\rho$, $P_{ij}(\rho)$, is the probability that state at $(t+1)$ is $j$ given that the state at $t$ is $i$ and the control input is $\rho$. That is, $P_{ij}(x) = P\{X(t+1) = j \mid X(t) = i, \rho\}$.

The new matrix, $(I - Q)^{-1}$, is called the fundamental matrix, $\textbf{\textit{M}}$ (Bhat 1972). From Equations (4.9) and (4.10), it is known that the $(i,j)$-th element of the fundamental matrix, $\mu_{ij}$, means the average number of visits to transient state $j$ starting from state $i$ before the process enters *any* absorbing state. We use the fundamental matrix to compute the probabilities and payoffs of $S$ and $F$ as follows.

Let $f_{ij}$ be the probability that the process starting in transient state $i$ ends up in absorbing state $j$. If we let the initial state be state 3, the Success state be state 2, and the Failure state be state 1, as in Figure 4, the probabilities of reaching $S$ and $F$ are $f_{3,2}$ and $f_{3,1}$, respectively.

Starting from state $i$, the process enters absorbing state $j$ in one or more steps. If the transition happens in a single step, the probability $f_{ij}$ is $P_{ij}$. Otherwise, the process may move either to another absorbing state (in which case it is impossible to reach $j$), or to a transient state $k$. In the latter case, we have $f_{kj}$. Hence,

$f_{ij} = P_{ij} + \Sigma_{k \in T} P_{ik} \cdot f_{kj}$ ,

which can be written in matrix form as

$F = R + Q \cdot F$,

and thus

$F = (I - Q)^{-1} \cdot R = M \cdot R. \tag{4.11}$

Therefore, the probabilities of $S$ and $F$ ($f_{3,2}$ and $f_{3,1}$) can be computed using the fundamental matrix ($\textbf{\textit{M}}$) and the sub-matrix ($\textbf{\textit{R}}$) of the original transition probability matrix.

Figure 6 depicts the fundamental matrix and the $\textbf{\textit{F}}$ matrix of the example MC in Figure 4. From the fundamental matrix, we know that when starting from state 3, the process visits state 3 once and state 4 about 0.4 times on average before it ends up in any absorbing state (i.e., $\mu_{3,3} = 1$ and $\mu_{3,4} = 0.4$). From the $\textbf{\textit{F}}$ matrix, we conclude that the probability of failure when starting from state 3 is 0.16 ($f_{3,1} = 0.16$), and the probability of success is 0.84 ($f_{3,2} = 0.84$).

$$\textbf{M} = \begin{bmatrix} 1 & 0.4 \\ 0 & 2 \end{bmatrix} \qquad\qquad \textbf{F} = \textbf{MR} = \begin{bmatrix} 0.16 & 0.84 \\ 0.8 & 0.2 \end{bmatrix}$$

**Figure 6: The fundamental matrix ($\textbf{\textit{M}}$) and the $\textbf{\textit{F}}$ matrix for the example MC.**

To compute the payoffs of $S$ and $F$, on the other hand, we need to compute the time discount, $TD(\Delta_{S/F})$. Let $\omega_{ij}$ represent a reward associated with each transition $i \rightarrow j$. Then, the reward for each transition represents the constant cost of delay, $c$. That is, $\omega_{ij} = c$, for all $i$, $j$ except $\omega_{0,0}$ and $\omega_{1,1}$. For both absorbing cases, the reward is 0.

We need to compute $\mu_{3i}^{(S)}$ and $P_{ij}^{(S)}$, where $\mu_{3i}^{(S)}$ is the number of visits to state $i$ starting from initial state 3 before the process enters $S$; and $P_{ij}^{(S)}$ is the conditional transition probability when the process ends up in $S$. Similarly, we need $\mu_{3i}^{(F)}$ and $P_{ij}^{(F)}$, where $\mu_{3i}^{(F)}$ is the number of visits to state $i$ starting from initial state 3 before the process enters $F$; and $P_{ij}^{(F)}$ is the conditional transition probability when the process ends up in $F$.

Those values can be computed by creating two new Markov chains ($\textbf{\textit{P}}^{(S)}$ and $\textbf{\textit{P}}^{(F)}$) from the original matrix $\textbf{\textit{P}}$, each of which has one absorbing state, $S$ and $F$, respectively. From $\textbf{\textit{P}}^{(S)}$, we can obtain $P_{ij}^{(S)}$ and compute the new fundamental matrix $\textbf{\textit{M}}^{(s)}$ (and therefore $\mu_{3i}^{(S)}$).

The new transition probabilities, $P_{ij}^{(S)}$, are the conditional probabilities that the process goes to state $j$ from state $i$ when the process ends up in $S$. Let $\phi$ be the statement "the original MC ends up in state $S$". Then,

$$P_{ij}^{(S)} = P(i \rightarrow j \mid \phi) = \frac{P((i \rightarrow j) \wedge \phi)}{P(\phi)} = \frac{P(\phi \mid i \rightarrow j) \cdot P(i \rightarrow j)}{P(\phi)} = \frac{f_{jS} \cdot P_{ij}}{f_{iS}} \qquad (4.12)$$

The new MC with the single absorbing state $S$, $\boldsymbol{P^{(S)}}$, is defined as follows.

$$\boldsymbol{P^{(S)}} = \begin{bmatrix} 1 & \boldsymbol{O} \\ \boldsymbol{R^{(S)}} & \boldsymbol{Q^{(S)}} \end{bmatrix},$$

where $\boldsymbol{R^{(S)}}$ is a column vector with $\boldsymbol{R^{(S)}} = \left\{\frac{P_{iS}}{f_{iS}}\right\}$, and $\boldsymbol{Q^{(S)}}$ is the matrix with

$$\boldsymbol{Q^{(S)}} = \left\{P_{ij}^{(S)}\right\} = \left\{\frac{f_{jS} \cdot P_{ij}}{f_{iS}}\right\}.$$

From $\boldsymbol{P^{(S)}}$, we can compute the new fundamental matrix $\boldsymbol{M^{(S)}}$, and therefore, $\mu_{3i}^{(S)}$. Of course, $P_{ij}^{(F)}$ and $\mu_{3i}^{(F)}$ are computed in a similar way.

Figure 7 depicts the two new Markov chains generated from the example MC. The new transition probabilities are computed using the conditional probability computation of Equation (4.12). To end up in failure, for example, the process always goes to state 4 from initial state 3 (if not, the process will end up in success). Therefore, the transition probability from state 3 to state 4 is 1 ($P_{3,4}^{(F)} = 1$) for the MC with the $F$ state.

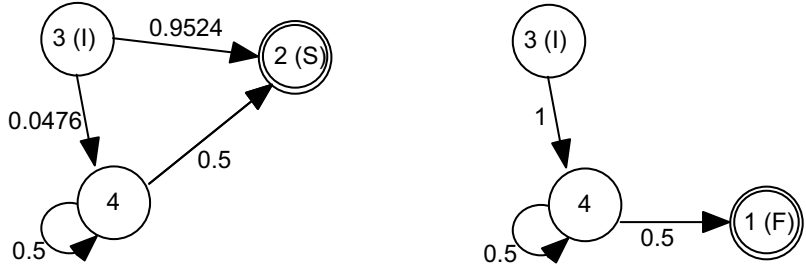

**Figure 7: Two new Markov chains with one absorbing state each, $S$ and $F$, respectively.**

$\Sigma_j P_{ij}^{(S)} \cdot \omega_{ij}$ is the average reward of the one-step state transition from state $i$ when the process ends up in $S$. Multiplying it by $\mu_{3i}^{(S)}$ (the number of visits to state $i$ starting from state $3$ until it goes to $S$), we compute the one-step reward accrued from state $i$ when the process ends up in $S$ ($\mu_{3i}^{(S)} \cdot \Sigma_j P_{ij}^{(S)} \cdot \omega_{ij}$). Adding this value for every state $i$, we compute the total reward of $S$. That is, the total reward of $S$ ($TD(\Delta_S)$) can be computed as follows, where $T$ denotes the set of transient states and $T^C$ denotes the set of absorbing states ($S$ and $F$).

$$TD(\Delta_S) = \sum_{i \in T} \left( \mu_{3i}^{(S)} \cdot \sum_{j \in \{T, T^C\}} P_{ij}^{(S)} \cdot \omega_{ij} \right). \qquad (4.13)$$

The reward of F ($TD(\Delta_F)$) can be computed in a similar way.

$$TD(\Delta_F) = \sum_{i \in T} \left( \mu_{3i}^{(F)} \cdot \sum_{j \in \{T, T^C\}} P_{ij}^{(F)} \cdot \omega_{ij} \right). \qquad (4.14)$$

Suppose the cost of delay is 0.1 (i.e., $\omega_{ij} = -0.1$). Then, $TD(\Delta_S)$ is 0.10952, and $TD(\Delta_F)$ is 0.3. If $U(Payoff(\rho))$ equals $Payoff(\rho)$ (i.e., risk neutral), the expected utility of the offer price of 5 (with its cost of 3) is 1.54 as computed in the following (using Equation (3.1)).

$u(5) = 0.84 \times (5 - 3 - 0.10952) + 0.16 \times -0.3 = 1.54.$

## 5. Evaluation

So far, we have argued for adding techniques based on Markov chain models into the AI repertoire to support intelligent decision-making (bidding) in a continuous double auction, and have described the techniques we have developed along with the rationale for particular modeling decisions we have made along the way. We now turn to the question of whether our new techniques, embodied in the p-strategy, indeed lead to improved decisions, and if so under what conditions. In essence, we are trying to answer the question of "*which bidding strategy should you, as a seller agent, use to maximize your profit?*," by comparing the p-strategy to some heuristic strategies across a range of situations. We also analyze how close (or far) the p-strategy is from performing as well as the *ex-post* optimum.

### 5.1. Experimental Testbed

As our approach to demonstrating the advantage of the p-strategy is through experiments, we need a systematic way of exploring a well-defined portion of the situations the p-strategy agent may potentially encounter, and a structured way of comparing its performance against other strategies. We have identified several experimental parameters that cover the possible auction situations (which we call the experimental space), and have developed a testbed that supports the identified parameters. Thus, by varying the value combinations of the experimental parameters of the testbed, we can systematically evaluate the performance of the p-strategy in different auction situations. (The testbed is available upon request from the first author.)

First, the testbed supports differing valuations [costs] of buyers [sellers], which reflect differences in their tastes [technologies]. The testbed supports uniform probability distributions. The valuation [cost] of buyer [seller] $i$ is drawn randomly from a uniform probability distribution $F_i$. In our experiments, we adjust the interval of the uniform distribution to be wide, medium, and narrow (*W, M, N*, respectively). Wide is 0 to 100, medium is 0 to 50, and narrow is 0 to 20.

Second, the testbed supports different degrees of dynamics. The dynamics of an auction are determined by two parameters: the arrival rate of buy-bids and the arrival rate of sell-asks. The number of buy-bids [sell-offers] and the interval between successive bids [offers] determines the arrival rate of buy-bids [sell-offers]. An agent's next bid [offer] arrives at an auction randomly between 0 and *offer-interval* since its last bid. So, the actual interval between bids of agent $i$ is (*offer-interval_i/2*) on average. The overall arrival rate of buyers and sellers can be computed by adding the arrival rates of individual agents as follows.

$arrival\,rate_{buyer} = \sum \frac{1}{(offer\_interval_i / 2)}$ , where $i \in buyers,$

$arrival\,rate_{seller} = \sum \frac{1}{(offer\_interval_i / 2)}$ , where $i \in sellers.$

For the dynamics of the auction, we vary the arrival rates of buy offers and sell offers to be either Low (*L*) or High (*H*). Low is 0.1, and high is 0.4. We use 10 buyers and 10 sellers, and adjust their *offer-interval*s accordingly. The *offer-interval* is 200 for *L*, and 50 for *H* arrival rates. By using different offer rates and different numbers of agents, we can simulate how active the auction is. As the MC model does not model individual agents, the number of agents in the system will not increase the complexity of the p-strategy reasoning.

Third, the testbed supports various agent populations. The demography of the agent population is determined by setting the bidding strategy of each individual agent $i$. The testbed provides several types of heuristic bidding strategies (in addition to the p-strategy).

As the auction places no restriction on the agent bidding strategy, many different types of agent strategies are possible. We characterize the space of agent strategies based on the amount of knowledge they use (see Table 2). The simplest is the strategy that requires no knowledge about the outside world. Examples include Zero-Intelligence (ZI) agents (Gode & Sunder, 1993) and Fixed-Markup (FM) and Random-Markup (RM) agents, both of which are described in the following. On the other hand, an agent may use a limited set of knowledge; it may use information about the status of the auction, how other agents behave, and the limited depth of other agents' interior reasoning processes. Of course, the more knowledge it uses, the more complex the strategy becomes. The Clearing-Price (CP) agent, described in the following, uses only information about the status of the auction (the clearing price). Most CDAs publish the clearing price (the highest standing buy-bid in our setting). Our p-strategy agent uses information about both auction status and participants' behavior.

The Recursive Modeling Method (RMM) agent uses models of the participants' internal reasoning in addition to information about the auction status (Gmytrasiewicz & Durfee, 2000). At the extreme, an agent may use all the information relevant to its decision making (e.g., a game-theoretic (GT) agent highlighted by Brandenburger & Dekel, 1993). We do not consider agent strategies that use information about the interior reasoning of other agents, as no GT strategy is known for the CDA setting and the RMM agent for the CDA is prohibitively expensive to build and run. Instead, we have developed three simple heuristic strategies, FM, RM, and CP, corresponding to the first two rows in Table 2.

We should note that several new agents have been developed specifically for CDAs (He et al., 2002; Tesauro & Das, 2001) since we ran the experiments reported here. Detailed comparisons with these new agents could be valuable, and this is an opportunity for future research, but would involve substantial efforts, in part because the experimental setup of the CDAs are somewhat different despite the fact that many common aspects exist. In this paper, we show based on the *post facto* analysis that there is room for improvement beyond our p-strategy, but it remains to be seen if other CDA strategies can do better.

|  | Auction | Participants | | Example strategies |
|---|---|---|---|---|
|  |  | Behavior | Interior reasoning |  |
| No information |  |  |  | ZI, FM, RM |
| Limited information | √ |  |  | CP |
| Limited information | √ | √ |  | P |
| Limited information | √ | √ | √ | RMM |
| All information | √ | √ | √ | GT |

**Table 2: The space of agent strategies.**

- **Fixed-Markup Strategy (FM-seller)**

The FM-seller bids its cost plus some predefined markup. The FM-strategy does not try to maximize the number of matches nor the profit per match. Rather, it is a satisficing strategy that hopes to gain a fixed profit whenever a deal is made. This is one of the simplest strategies because the seller does not have to monitor the auction nor build a model of other agents.

- **Random-Markup Strategy (RM-seller)**

The RM-seller bids its cost plus some random markup. The RM-seller is a "budget-constrained zero-intelligence trader" who generates random bids subject to a no-loss constraint (Gode & Sunder, 1993).

- **Clearing-Price Strategy (CP-seller)**

The CP-seller receives information about the next clearing price (clearing-price quote) from the auction agent and submits the quoted clearing-price as its offer as long as it is higher than its cost. When its cost is higher than the quoted clearing price, it behaves like the FM-strategy (i.e., bids its cost plus some fixed markup).

The CP strategy is an optimal strategy at the current snapshot of the auction provided no other agent arrives at the auction before its bid, because the quoted clearing price is the highest profit achievable by the next incoming seller. The actual clearing price, however, will change if new offer(s) arrive during the time between the clearing-price quote and the CP-seller's offer.

Intuitively, the clearing-price strategy seems a good heuristic when the auction is less dynamic. By ignoring the dynamics of the auction (i.e., incoming buyers and sellers), however, the CP-seller cannot capitalize on future, more lucrative deals, and may miss out on the current deal due to other incoming seller(s) with lower sell price(s). In comparison, the p-strategy does anticipate the future by modeling the auction process stochastically, and therefore can weigh the tradeoffs between the current known deal and the more lucrative (but uncertain) deals in the future.

- **Post-facto Optimal Strategy (OPT-seller)**

In addition, we have developed an optimal bidding strategy (called OPT-strategy). No matter how many strategies the p-strategy is compared against and how well it performs, it will not prove that the p-strategy is the best; we will always wonder whether there exists a better strategy. Instead of exhaustively comparing the p-strategy against every possible strategy (which is impossible), we measure the maximum profit an agent may have achieved had it known exactly how the future would unfold. The post-facto optimal profit is not achievable in the real world since it is based on a post-facto analysis of what had happened in the auction, but it serves as a benchmark of what the highest profit might have been.

The main idea in the post-facto optimal strategy is to find the best offer at time $x$ as if the agent knows about the future auction process with a probability of 1. The OPT-strategy analyzes the auction process after the auction has run and discovers the best offer price. That is, it can look ahead to the time before it makes another offer, and finds out the best price for the current offer.

As the main purpose of our experiments is to evaluate the performance of the *seller's* strategy, we do not test different buyer's strategies; we let all buyers be competitive (i.e., they bid their true valuations). Sellers, on the other hand, can use different strategies (FM, RM, CP, OPT, and p-strategy). The resulting demography of agent populations can be from all sellers' using the FM-strategy to all sellers' using the p-strategy. That is, the number of agents using the fixed-markup, the random-markup, the clearing-price-bid, and the p-strategy can vary from *(0,0,0,0)* to *(10,0,0,0)* to *(0,10,0,0)* to *(0,0,10,0)* to *(0,0,0,10)* for each experiment. Let $n$ be the number of agents and $s$ be the number of strategies, then the number of possible agent populations can be $O(n^5)$. It is indeed a huge experimental space, and we test some notable cases in the following section.

Note that since the p-strategy agent does not model individual agents, the number of agents in the auction is less significant in the experimental results. By changing the offer rates and by changing the population of 10 sellers, the experimental testbed can simulate various environments the p-strategy agent may face, such as a large population of active participants to a small population of less active participants. Note that one could have simulated the dynamics of the auction without having 10 competing agents. A single agent with varying offer rates would simulate the dynamics. The reason for having multiple agents in the system is for simulating different population demographies.

A typical experiment presented in this paper consists of 20,000 cycles, unless noted otherwise. The rest of the section presents a subset of notable cases from an extensive suite of experiments (Park 1999). The choices of parameters reported in this paper are based on information gleaned from those experiments. For example, we selected 0.1 and 0.4 to represent low or high arrival rates (and thus represent the degree of activity in auctions.) The markups of 5 and 25 presented in this paper represent small and large markups giving qualitatively different results in the extensive experiments. That is, from the experiments that find the optimal markup in different auction environments, we have found different "optimal" markup values (near 5 for narrow zone and near 25 for medium or wide zone). Thus we use these to make the other agents as competitive as possible. The experiments with markups near 5 or 25 show trends similar to what is reported here.

## 5.2. Comparison of Agent Strategies

We compare the performance the p-strategy to that of the other strategies (FM, RM, CP, and OPT). The main questions are:

- Whether the p-strategy outperforms what we conceive as reasonable, realistic agent strategies (except the OPT-strategy), and
- How closely the p-strategy performs compared to the (ideal) OPT-strategy.

We vary the negotiation zones and offer arrival rates of buyers and sellers. To compare the agent strategies, we replace the target seller being compared with the FM, RM, CP, P, and OPT strategy sellers in each set of experiments, while letting all the other sellers participating in the auction simply bid their true costs. The markup of the FM and CP-strategy is set to five. Figure 8[5] depicts the profits of the FM, RM, CP, P, and OPT sellers (represented as F, R, C, P, O, respectively). The experiment in Figure 8-(a)-(ii) depicts the performance of the compared strategies in the setting where both buyers and sellers have narrow negotiation zones and the arrival rates of buy and sell offers are 0.1 and 0.4, respectively, for example.

The FM-seller achieves stable profit regardless of negotiation zones, because of its *satisficing* behavior of trying to gain the same fixed markup, while the profits of the other types of sellers change significantly (compare the FM-seller's profits, for example, in Figure 8-(a)). The FM-seller performs relatively well as compared to the other types of sellers when there are more sell offers (1_4 cases).[6] More competition among sellers means a lesser amount of potential gain, and thus the simple FM-strategy works well. In addition, the markup value of the FM-seller is five in our experiments, which happens to be close to the optimal markup value for the experiment in Figure 8-(b)-(ii) (which is 7), as based on our more complete experiments (Park 1999). As a result, with a lucky guess the FM-seller makes profit comparable to that of the p-seller. In general, however, the FM-seller works poorly compared to the other strategies because of its "static" fixed markup. In the case of wider negotiation zones or higher demand from buyers, in particular, the FM-seller fails to capture higher profit by insisting on the same (small) fixed markup (see the experiment in Figure 8-(c)-(iii), for example).

---

[5] Note that all the subsequent graphs in this paper have different vertical scales.

[6] We use the notation "$x\_y$ case" to denote the test set where the buyer's arrival rate is $0.x$ and the seller's arrival rate is $0.y$. So, the 1_4 case, for example, denotes the experiment where the buyer's arrival rate is 0.1 and the seller's arrival rate is 0.4.

The performance of the RM-strategy fluctuates widely across the experiments. In general, the RM-strategy performs worse than the other strategies. By randomly raising its bid, the RM-seller makes higher profit than the FM-seller when the negotiation zone is wider. This observation, however, does not hold when we select other markup values for the FM-seller. The FM-seller's markup of 5 is too small a profit to pursue in the wide negotiation zone, and the RM-seller performs always worse than the FM-seller with an optimal markup value as described in the following section. As the profit gain of the RM-seller is rather unpredictable and low, we conclude that randomizing the profit markup is a poor heuristic to use.

The CP-seller performs especially well when there are more buy offers than sell offers. Thus, the CP-strategy is almost as good as the p-strategy in experiments with the arrival rates of (0.4 & 0.1), as shown in Figure 8). We attribute this result to two reasons. First, more demands from buyers mean more opportunities for higher profit gains for sellers. Secondly, and more importantly, due to fewer sellers, the clearing price is less likely to change (*adversely* to the CP-seller).

The CP-seller achieves slightly higher profit than the p-seller in the experiment in Figure 8-(b)-(ii). Note that the advantage of the p-seller over the CP-seller comes from p-seller's bidding higher than the current clearing price to take advantage of more lucrative (future) incoming trades. With high competition, however, higher bidding results in fewer matches, and reduces the p-seller's profit. Therefore, the p-strategy is not very effective in the 1_4 case in general. In addition, the markup of 5 is close to the optimal value for the experiment in Figure 8-(b)-(ii) and the p-seller does not perform well in the 1_4 case. (Recall that the CP-seller submits the current clearing-price quote when it is higher than its cost, or its cost plus some fixed markup when its cost is higher than the clearing-price quote (i.e., behaves like the FM-seller).) Note that the markup of 5 is intentionally chosen to make the CP-seller the most competitive in the 1_4 case.

The CP strategy, however, does not perform well with a wide negotiation zone, because it may have achieved a higher profit had it waited for incoming buy offers with higher price rather than settling at the current clearing price. That is, the same reason that makes the CP-seller successful in the 1_4 case acts against it in the wide negotiation zone case.

Table 3 shows the results of a t-test, where each agent strategy (FM, RM, and CP) is compared to the p-strategy under different negotiation zones and different arrival rates. The t-test is useful for comparing two samples with a certain confidence level. In our case, it is used to analyze where one strategy is better than the other. It not only gives a yes-no answer but also gives a confidence level on the answer. Knowing the confidence level is often more helpful to agent designers than the simple yes-no answer from the hypothesis tests.

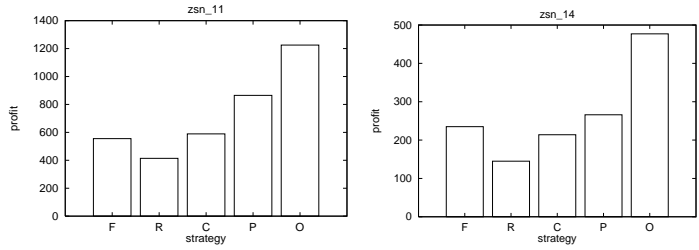

(i) Arr. rates of (0.1 & 0.1)    (ii) Arr. rates of (0.1 & 0.4)

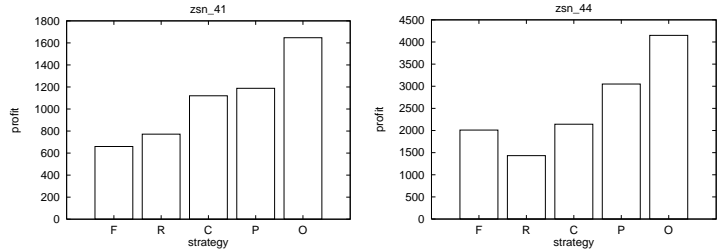

(iii) Arr. rates of (0.4 & 0.1)    (iv) Arr. rates of (0.4 & 0.4)

(a) When the negotiation zone is narrow

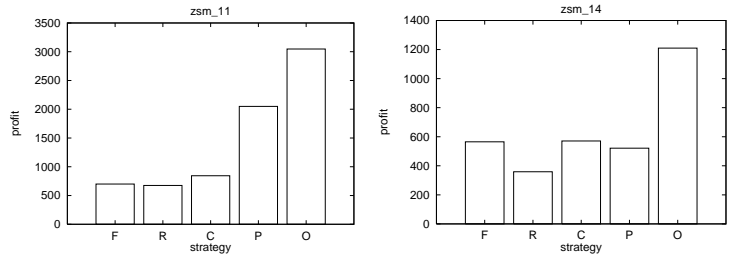

(i) Arr. rates of (0.1 & 0.1)    (ii) Arr. rates of (0.1 & 0.4)

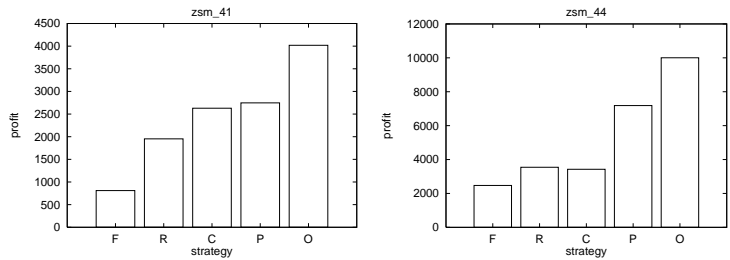

(iii) Arr. rates of (0.4 & 0.1)    (iv) Arr. rates of (0.4 & 0.4)

**(b) When the negotiation zone is medium**
**Figure 8: The profit of the agent strategies (continued on the next page).**

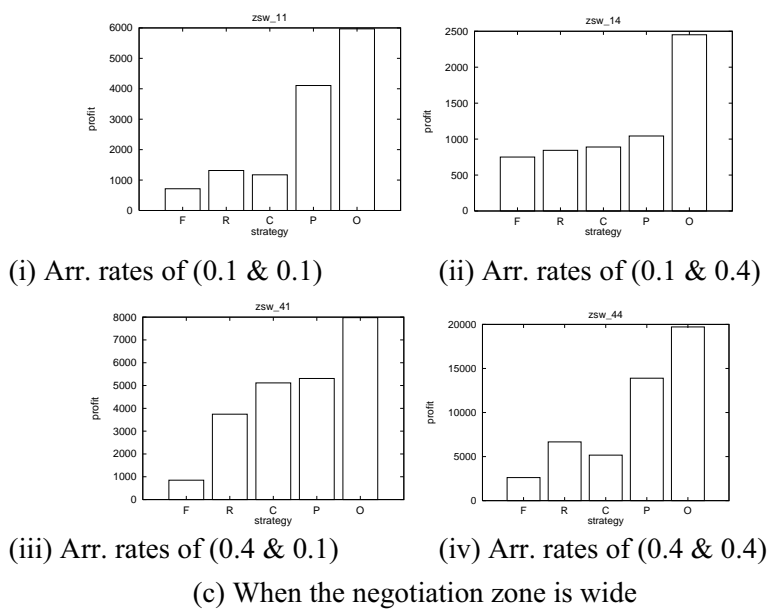

(i) Arr. rates of (0.1 & 0.1)      (ii) Arr. rates of (0.1 & 0.4)

(iii) Arr. rates of (0.4 & 0.1)     (iv) Arr. rates of (0.4 & 0.4)

(c) When the negotiation zone is wide

**Figure 8: The profits of agent strategies (continued from the previous page).**

Each cell in Table 3, representing a t-test of 20,000 data points, lists the better strategy of the two compared in each environment, given the confidence level of 90%. For each t-test, we compute the two sample means (of the p-strategy and the other strategy compared), the sample standard deviations, the mean difference, the standard deviation of the mean difference, the effective number of degrees of freedom, and the confidence interval for mean difference. If the confidence level includes zero, the difference is not significant at the 90% confidence level. If the confidence interval does not include zero, the sign of the mean difference indicates which strategy is better. For more detailed information about the t-tests performed, one may refer to (Park, 1999).

| Arrival rates | 0.1 & 0.1 | 0.1 & 0.4 | 0.4 & 0.1 | 0.4 & 0.4 |
|---|---|---|---|---|
| FM | P | P | P | P |
| RM | P | P | P | P |
| CP | P | P | ? | P |

(a)  When the negotiation zone is narrow

| Arrival rates | 0.1 & 0.1 | 0.1 & 0.4 | 0.4 & 0.1 | 0.4 & 0.4 |
|---|---|---|---|---|
| FM | P | ? | P | P |
| RM | P | P | P | P |
| CP | P | ? | ? | P |

(b)  When the negotiation zone is medium

| Arrival rates | 0.1 & 0.1 | 0.1 & 0.4 | 0.4 & 0.1 | 0.4 & 0.4 |
|---|---|---|---|---|
| FM | P | P | P | P |
| RM | P | P | P | P |
| CP | P | P | ? | P |

(c)  When the negotiation zone is wide

**Table 3: The results of t-test.**

The p-seller always outperforms the RM-seller. As we noted earlier, randomly raising its bid price is not a good heuristic. Compared to the FM and CP sellers, the p-seller performs better when buy and sell offers arrive at a similar rate (i.e., either 1_1 or 4_4 cases). We cannot conclude, however, which strategy is better when the negotiation zone is medium and the arrival rates are (0.4 & 0.1), given the confidence level of 90%. We also find out the CP-strategy performs as well as the p-strategy in the 4_1 cases regardless of the size of negotiation zone. As discussed in the previous section, the CP-strategy is indeed a good heuristic to use in the 4_1 cases.

No agent can have a complete, deterministic view of the current and future status of the auction except the OPT-seller, but the agent that takes those uncertainties into account should have an advantage over those who do not. That is why the p-strategy that models the auction process is a better strategy in general.

### 5.3. Optimal Markups of the FM and CP Strategies

The markup value of 5 is close to the optimal markup value for both the FM and CP sellers in the experiment in Figure 8-(b)-(ii), which contributes to their high performance (i.e., not significantly different from the p-seller) in that experiment. This raises the question of whether the FM-seller with an optimal markup will be better than (or similar to) the p-seller in other cases as well. Thus, we have investigated the relation between the markups and the performance of the FM-seller and the CP-seller.

To answer this question, we first find the optimal markup of the FM and CP strategies in each experimental setting. After determining the optimal markup, we then run the same experiments in the previous section again, this time using the FM and CP sellers with an optimal markup.

From the experiments (not shown), we determine the best markup value for each case. As expected, the optimal markup value of the FM strategy increases (1) when the negotiation zone is wide, and (2) when there are more buy offers. The optimal value decreases, on the other hand, when the competition among sellers increases. Similar to the case of the FM-seller, the optimal markup of the CP strategy is higher with a wider negotiation zone, and with more buy offers. However, compared to the wide difference of the FM-seller's profit depending on different markup values, the profit of the CP-seller is less sensitive to the markup value, because the CP-seller uses the markup value only when its cost is higher than the current clearing-price quote.

When comparing the p-seller to the FM and CP sellers with an optimal markup value (called opt-FM and opt-CP, respectively), we find that the profits of the opt-FM and opt-CP sellers are higher than the profits of the FM and CP sellers with a markup of 5 (compare with Figure 8). In particular, the opt-FM and opt-CP sellers perform well in the 1_4 cases regardless of the negotiation zone size

The t-test results in Table 4 indicate that the opt-FM and opt-CP are actually better than the p-strategy when the buy and sell arrival rates are 0.1 and 0.4, respectively, and the negotiation zone is medium, and as good as the p-strategy in other 1_4 cases. Similar to previous results, the opt-CP continues to perform well with high demand from buyers (i.e., the 4_1 case). In other cases, the p-strategy is still a better strategy.

Using a pre-determined markup value, however, assumes that an agent knows what an optimal markup for the current situation is. This assumption is not generally reasonable for the following two reasons. First, the possible auction situations are too large to exhaustively search for the optimal markup value. Secondly, in the dynamic CDA, the optimal markup may vary with time. In fact, finding an optimal markup for the current situation is what the p-strategy is trying to do by modeling the auction process stochastically and adjusting its offer price accordingly.

| Arrival rates | 0.1 & 0.1 | 0.1 & 0.4 | 0.4 & 0.1 | 0.4 & 0.4 |
|---------------|-----------|-----------|-----------|-----------|
| Opt-FM | P | ? | P | P |
| Opt-CP | P | ? | ? | P |

(a) When the negotiation zone is narrow

| Arrival rates | 0.1 & 0.1 | 0.1 & 0.4 | 0.4 & 0.1 | 0.4 & 0.4 |
|---------------|-----------|-----------|-----------|-----------|
| Opt-FM | P | Opt-FM | P | P |
| Opt-CP | P | Opt-CP | ? | P |

(b) When the negotiation zone is medium

| Arrival rates | 0.1 & 0.1 | 0.1 & 0.4 | 0.4 & 0.1 | 0.4 & 0.4 |
|---------------|-----------|-----------|-----------|-----------|
| Opt-FM | P | ? | P | P |
| Opt-CP | P | ? | ? | P |

(c) When the negotiation zone is wide

**Table 4: The t-test results (comparing the p-strategy to the opt-FM and opt-CP strategies).**

## 5.4. Comparison to the OPT-strategy

Instead of comparing the p-strategy to all the feasible agent strategies (which is impossible), we compare the p-strategy to the OPT-strategy (i.e., the ideal upper bound). In particular, we are interested in how closely the p-strategy performs to the OPT-strategy. Figure 9 shows the profit per offer of the agent strategies normalized to that of the OPT-strategy.

Note that the p-seller adjusts its offer price depending on the auction situation. When the negotiation zone is narrow, for example, it tries to increase the number of matches. When the negotiation zone is wide, it tries to increase the profit per match. When the arrival rate of buy offers is high, it tries to increase the profit per match, while it tries to increase the number of matches when the arrival rate of sell offers is high. Similar behavior is also observed from the OPT-seller, while the FM, RM, and CP-sellers show totally different behaviors. Given the resemblance to the behavior of the OPT-strategy, it is not surprising that the p-strategy performs the best overall. When there is high competition among sellers, however, the p-strategy performs similarly to the CP-strategy.

Two observations are derived by comparing the agent strategies to the OPT-strategy. First, the p-strategy performs most closely to the OPT-strategy in most cases, and in Figure 9-(a) and (d), the p-strategy is clearly the winning strategy among those compared. When buy and sell offers arrive at a similar rate, and when other sellers behave naïvely (i.e., the other nine sellers bid their true costs), the p-strategy always outperforms the other agent strategies. When there is higher demand from buyers, both the CP-seller and the p-seller perform similarly well (see Figure 9-(c)).

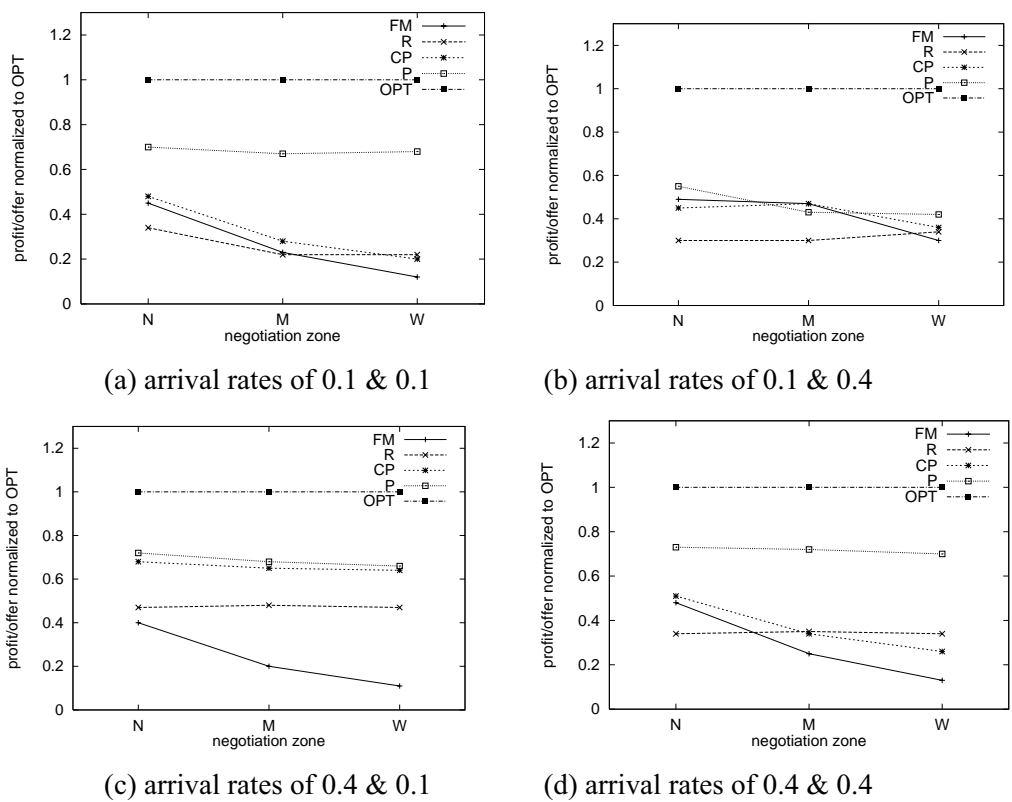

(a) arrival rates of 0.1 & 0.1

(b) arrival rates of 0.1 & 0.4

(c) arrival rates of 0.4 & 0.1

(d) arrival rates of 0.4 & 0.4

**Figure 9: Comparison of the profit per offer normalized to OPT.**

Secondly, the p-seller's performance degrades with high competition, and the FM or CP strategies can get similar profits to that of the p-strategy in such cases (most notably, the Medium negotiation zone case in Figure 9-(b)). The difference between the OPT-seller and the p-seller is most significant with high competition among sellers. With high competition, the sellers have little room for profit gain to begin with. In addition, the background sellers who bid their true costs gain lots of matches, which leaves even less room for strategic bidding. The p-strategy of trying to achieve high profit per match (by looking into the future of the auction process) is not very effective in this situation, and the difference between a simpler strategy and a more sophisticated strategy (like the p-strategy) is smaller.

Using the OPT-strategy's behavior to improve the p-strategy seems like a good extension to the p-strategy. Since the OPT-strategy uses information unattainable in the real setting (such as the future clearing price), however, we are still investigating what could be usable. More generally, the OPT-strategy provides a sense of how well our p-strategy does compared to what is theoretically possible, which overall is quite well, but suggests that there is room for improvement.

## 5.5. Comparison of Agent Strategies under Different Agent Populations

In the experiments presented so far, we have varied only the negotiation zones and offer arrival rates, and the agent population has been fixed such that all the other (background) sellers in the auction bid their true costs (i.e., FM-sellers with markup = 0). The types of background sellers, however, are another factor (in addition to the negotiation zones and offer arrival rates) that influences the performance of an agent strategy.

In this section, we try various types of background sellers, and examine the performance of the p-strategy under different agent populations. Specifically, we want to find out whether the p-strategy continues to outperform the other agent strategies even under different agent populations.

Figure 10 depicts the possible configuration of background sellers. The *x, y,* and *z* axis corresponds to the number of FM, CP, and P background sellers (i.e., sellers except the target agent being compared), respectively. In the experiments with nine background sellers, any integer point on the plane connecting (9, 0, 0), (0, 9, 0), and (0, 0, 9) is a possible configuration of seller population. The previous experiments with 9 sellers who bid their true costs, for example, correspond to the point (9, 0, 0) on the *x*-axis.

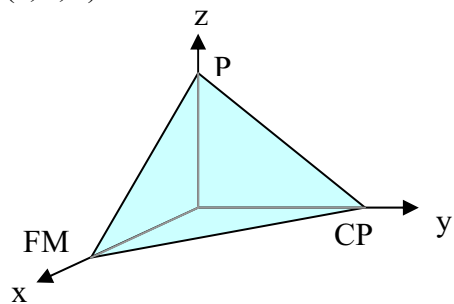

**Figure 10: The space of different agent population.**

To examine the performance of the p-strategy under different agent populations, we now use different background sellers. We still keep the fact that there is only one p-seller in the auction for now. That is, we examine some notable cases on the line connecting (9, 0, 0) and (0, 9, 0). (The cases where there are more than one p-strategy seller are examined in the next section.) They are:

- All the background sellers bid their costs plus 25[7] (i.e., FM sellers with markup 25),
- All the background sellers use the CP-strategy with markup 0,
- Some sellers use the CP-strategy with markup 0 and others use the FM-strategy with markup 0.

Figure 11 shows the experimental results of the cases when all the background sellers are FM-sellers with markup 25. As we are only interested in general trends (and not the specific profit values), and as the trends are similar across different negotiation zones, we only show the medium negotiation zone.

The most notable difference as compared to the experiments with true-cost bidding background sellers is that all the compared agents now receive more profits. As the background FM-sellers with higher markup receive fewer matches than the background FM-sellers with markup 0, there are more opportunities for matches, and as a result, the profits of all the compared sellers increase. The change in the background sellers favors in particular the p-seller and the CP-seller, as they can achieve more matches and more profit per match. The change is most notable in the CP-strategy in the 4_1 case. Similar to the case where the background sellers bid their true costs, the p-strategy is still the best strategy when buy and sell offers arrive at a similar rate (the 1_1 and 4_4 cases).

---

[7] The FM-seller's optimal markup for medium or large negotiation zones is near 25. We intentionally set the markup to be 25 to make the FM-seller most competitive (Park 1999).

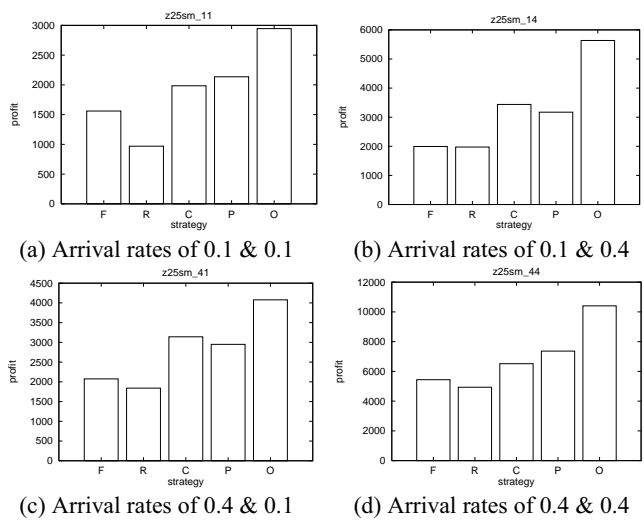

(a) Arrival rates of 0.1 & 0.1          (b) Arrival rates of 0.1 & 0.4

(c) Arrival rates of 0.4 & 0.1          (d) Arrival rates of 0.4 & 0.4

**Figure 11: Profit of the agent strategies (when the other sellers use the FM-strategy with a markup of 25).**

Figure 12 shows the cases when all the other sellers use the CP-strategy with a markup of 0. As the background CP-sellers receive more profits than the background FM-sellers, all the compared sellers receive less profit than in the previous experiments. Notice that the profit of the CP-strategy decreases the most. Now the CP-strategy cannot achieve higher profit than the p-strategy in any case, due to the contention among multiple CP-sellers. That is, the CP-strategy performs poorly when there are multiple CP-sellers, which is an important property of the CP-strategy.

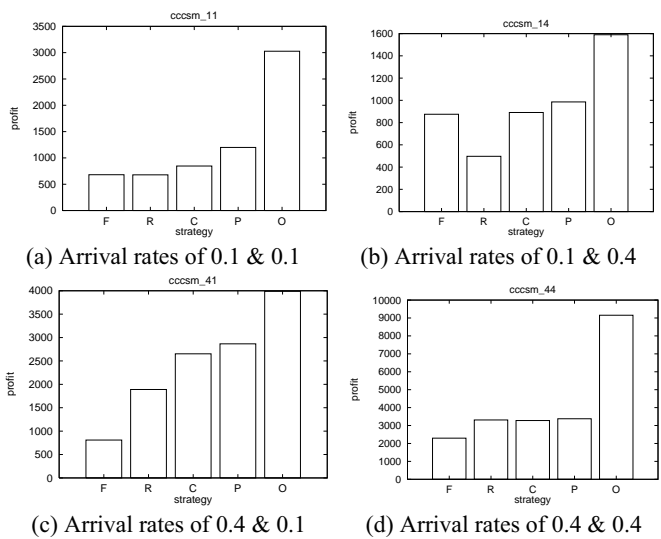

(a) Arrival rates of 0.1 & 0.1          (b) Arrival rates of 0.1 & 0.4

(c) Arrival rates of 0.4 & 0.1          (d) Arrival rates of 0.4 & 0.4

**Figure 12: Profit of the agent strategies (when the other sellers use the CP-strategy with a markup of 0).**

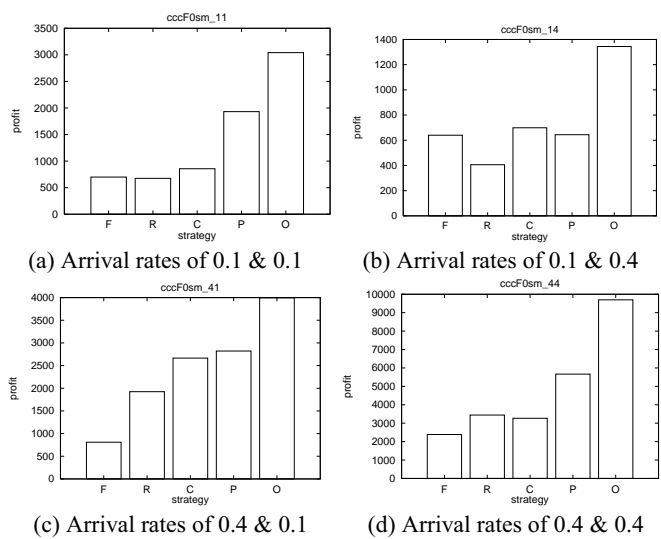

(a) Arrival rates of 0.1 & 0.1

(b) Arrival rates of 0.1 & 0.4

(c) Arrival rates of 0.4 & 0.1

(d) Arrival rates of 0.4 & 0.4

**Figure 13: Profit of the agent strategies (when 4 sellers use the FM-strategy with a markup of 0 and 5 sellers use the CP-strategy with a markup of 0).**

When the background sellers are comprised of both the FM and CP sellers, we see the p-strategy is the best strategy except in the 1_4 case, as usual. Figure 13 represents the cases where four sellers use the FM-strategy with a markup of 0 and five sellers use the CP-strategy with a markup of 0. Similar trends are reported in the cases where four sellers use the FM-strategy with a markup of 25 and five sellers use the CP-strategy with a markup of 25.

As the performance of an agent strategy depends on many factors happening in the auction, it is hard to draw a single sweeping conclusion. However, the experimental results show that (1) the p-strategy still performs the best in most cases, and that (2) the CP-seller's performance is most sensitive to the demography of agent population.

### 5.6.  Performance of the P-strategy in the Presence of Competing P-strategy Sellers

Given that the p-strategy is effective in the CDA in general (from the previous sections), nothing prohibits any self-interested seller from adopting the p-strategy, and thus we are interested in the collective behavior of the p-sellers when multiple p-sellers coexist in the system. We have investigated how the absolute and relative performance of a p-seller changes because of the presence multiple p-sellers.

In our experiments, we increase the number of background p-sellers while decreasing the number of the other background sellers in the auction. The types of other background agents used in the experiments are:

(1) FM-sellers with a markup of 5,

(2) FM-sellers with a markup of 25, and

(3) CP-sellers with a markup of 0.

The markup value of 5 instead of 0 is used to compare the profits of the p-seller and the FM-seller. A FM-seller with a markup of 0 does not accrue any profit, which makes it impossible to compare the relative performance of the FM- and p- sellers.

By changing the number of p-sellers in the auction, we answer two questions. First, does the p-seller's performance degrade with multiple p-sellers due to competition among p-sellers? Second, if the p-seller's profit decreases as we expect, does any other strategy perform better than the p-strategy in the presence of multiple p-sellers?

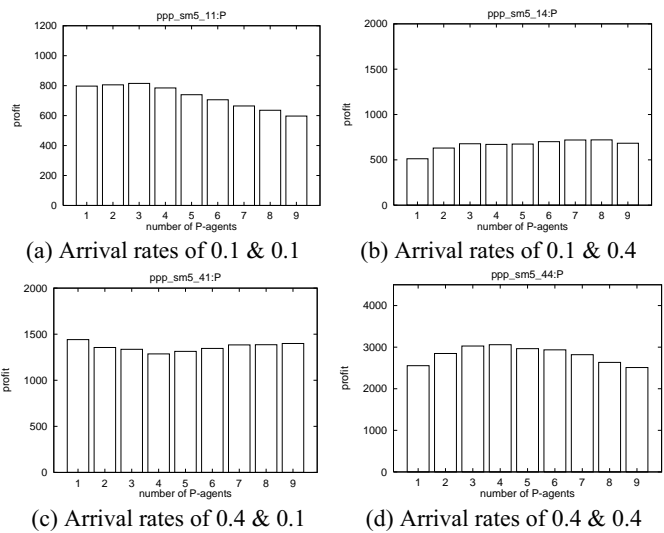

(a) Arrival rates of 0.1 & 0.1

(b) Arrival rates of 0.1 & 0.4

(c) Arrival rates of 0.4 & 0.1

(d) Arrival rates of 0.4 & 0.4

**Figure 14: The change of profit of the p-seller as the number of p-strategy sellers increases (when the FM-seller's markup is 5).**

Figure 14 shows the changes in the p-seller's profit as the number of p-sellers increases (and the number of FM-sellers decreases). The profit of the p-seller decreases with more p-sellers in (a) and (d) due to the competition among p-sellers.

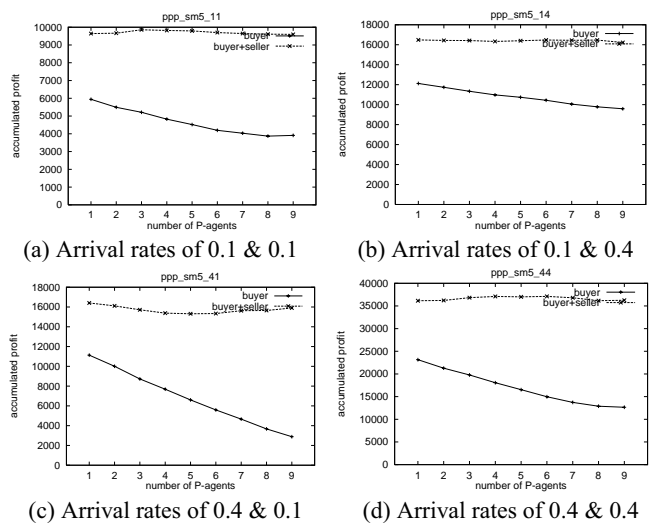

(a) Arrival rates of 0.1 & 0.1

(b) Arrival rates of 0.1 & 0.4

(c) Arrival rates of 0.4 & 0.1

(d) Arrival rates of 0.4 & 0.4

**Figure 15: Total profit generated in the auction (when the FM-seller's markup is 5).**

The profit of the p-seller, on the other hand, does not change significantly in the 1_4 and the 4_1 cases. These results can be explained by looking at the total profit generated in the entire auction system. Figure 15 depicts the total profits gained by all the buyers and all the sellers. As the number of FM-sellers with a low markup value of 5, who receive most of the matches, decreases, more buyers are available to the other sellers, and the p-sellers are able to take advantage of that. So, the p-sellers receive more matches in the 1_4 case, and they receive more profit per match by increasing their offer prices in the 4_1 case. That is, in both cases, instead of squeezing the profit out of competing p-sellers, the p-strategy sellers are able to gain similar profit at the expense of buyers.

Figure 16 shows the profit of the p-strategy seller when the FM-seller's markup is 25. In most cases (except the 4_1 case), the profit of the individual p-seller decreases as the number of p-sellers increases. With more p-sellers, the competition among the p-sellers increases, and they eat into each other's profits. In the 4_1 case, however, the profit of the p-seller does not decrease with the increase in p-sellers. With the availability of many buy offers, a p-seller extracts profits from the buyers rather than from the other p-sellers. Therefore, the number of matches does not decrease after a certain point. See Figure 17 for the changes of the buyers' and sellers' profits.

Now, another question is how does a simpler strategy seller perform in the presence of multiple p-strategy sellers? In particular, does it outperform the p-seller in the presence of multiple p-sellers? The FM-seller with a markup of 5 does not perform well against the p-sellers, and receives almost the same profits regardless of the number of p-sellers. The markup of 5 is too small for the medium negotiation zone, so the FM-seller is not able to gain profit close to that of the p-seller. The profit of the FM-seller with a markup of 25 is generally smaller than that of the p-seller, but the difference decreases with the increase of p-sellers. In particular, the FM-seller performs better than the p-seller when the number of p-sellers increases in the 4_4 case. The result indicates that a seller may want to switch between p-strategy and a simpler strategy depending on what the other sellers are doing. By dynamically switching to a simpler strategy, an agent can achieve a similar profit (to that of using the p-strategy) while exerting less effort (time and computation) on calculating bids.

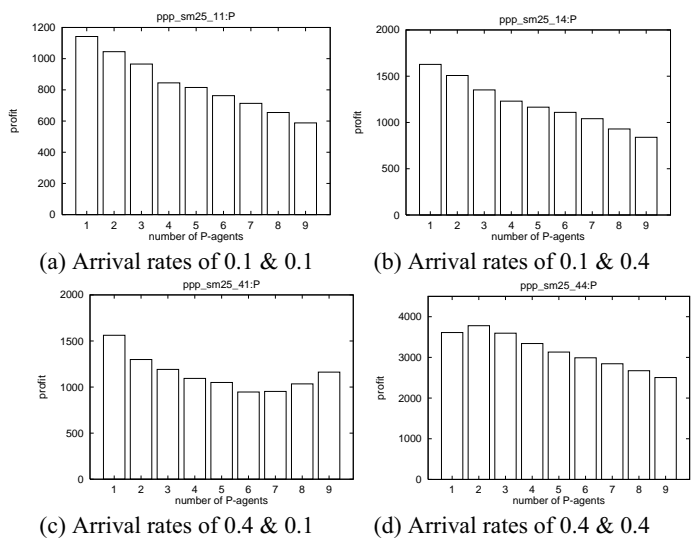

(a) Arrival rates of 0.1 & 0.1      (b) Arrival rates of 0.1 & 0.4

(c) Arrival rates of 0.4 & 0.1      (d) Arrival rates of 0.4 & 0.4

**Figure 16: The change of the profit of the p-seller as the number of p-strategy sellers increases (when the FM-seller's markup is 25).**

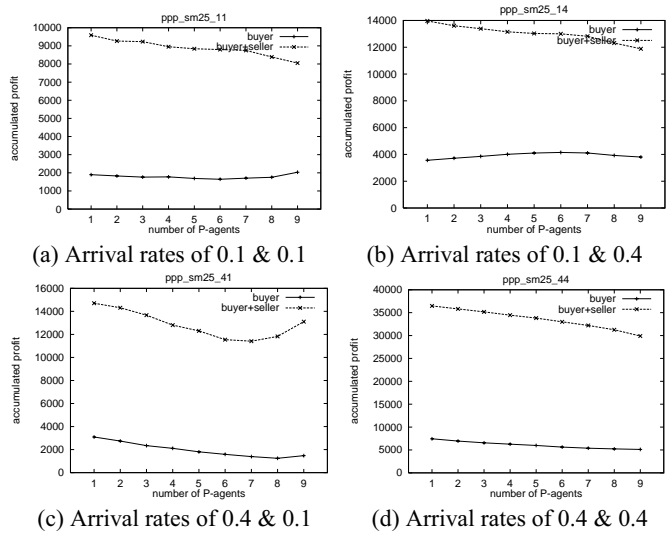

(a) Arrival rates of 0.1 & 0.1    (b) Arrival rates of 0.1 & 0.4

(c) Arrival rates of 0.4 & 0.1    (d) Arrival rates of 0.4 & 0.4

**Figure 17: Total profit generated in the auction (when the FM-seller's markup is 25).**

In the final set of experiments, we examine the performance of the p-seller when multiple p-sellers and CP-sellers coexist (in Figure 18). Interestingly, the p-seller's profit does not decrease with the increase of p-sellers. More surprisingly, the buyers' profit and the overall system's profit increase with an increase in p-sellers (as in Figure 19).

The reason for such different behavior (as compared to the cases with FM-sellers as background sellers) can be explained by two reasons. First, the CP-seller's performance is significantly degraded in the presence of the other CP-sellers. As shown in Figure 20, the profit of the CP-seller is lower when there are many CP-sellers due to contention among themselves. As the number of other CP-sellers decreases, the profit of a CP-seller increases, which results in the increase of total sellers' profit. Secondly, the p-seller's performance is not affected by multiple CP-sellers. Notice that the profit of the p-seller with multiple CP-sellers is less to begin with, as compared to the case with multiple FM-sellers.

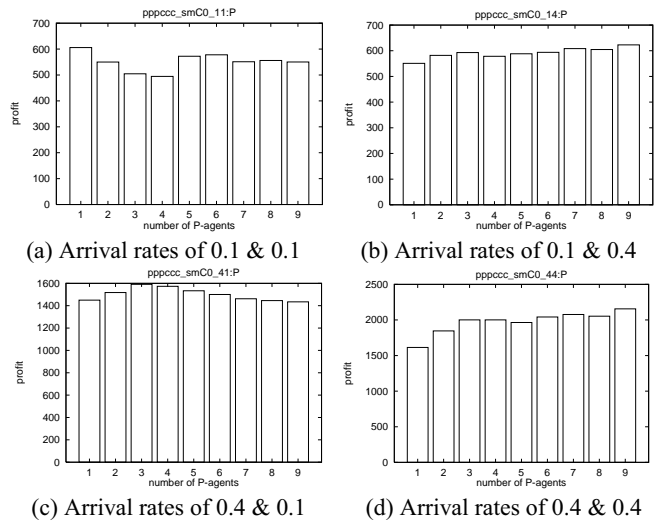

(a) Arrival rates of 0.1 & 0.1    (b) Arrival rates of 0.1 & 0.4

(c) Arrival rates of 0.4 & 0.1    (d) Arrival rates of 0.4 & 0.4

**Figure 18: The change of the profit of the p-seller as the number of p-sellers increases in the presence of multiple P and CP sellers.**

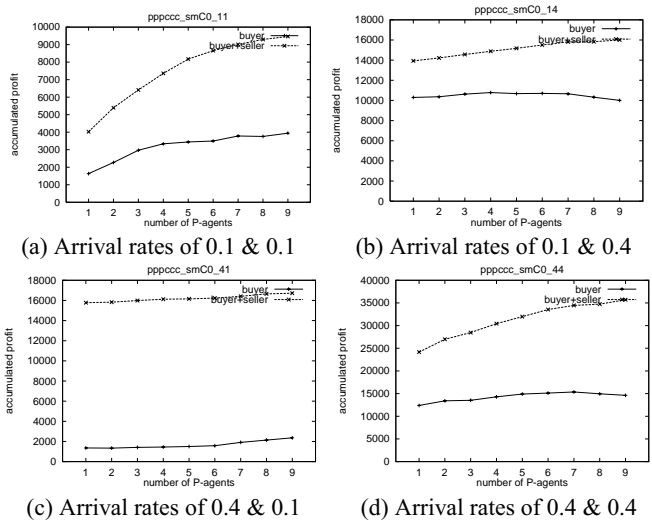

**Figure 19: Total system profit generated in the auction with multiple P and CP sellers.**

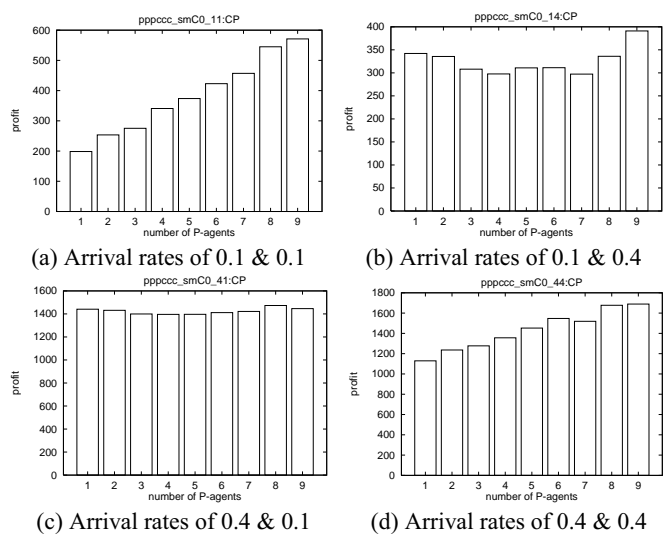

**Figure 20: Profits of the CP-seller in the presence of multiple P and CP sellers.**

Competition among p-sellers degrades the performance of each p-strategy seller, but the degree of performance degradation depends on the demography of the agent population. It is most significant when the background sellers use the FM-strategy with a high markup, and least significant when the background sellers use the CP-strategy.

From the view of total system profit (i.e., the sum of all the buyers' profit and all the sellers' profit), an increase in the number of p-sellers reduces the buyer-side profits and does not reduce the total system profit much. It is important to note that the p-strategy does not impair the efficiency of the auction much. In contrast, the CP-strategy significantly degrades the total system profit when there are multiple CP-strategy sellers by reducing the number of matches.

## 6. Conclusion

The p-strategy is based on stochastic modeling of the auction process. Ignoring that other agents behave strategically may sound like a risky approximation, but as the behavior of strategic agents is reflected in the dynamics of the CDA, the stochastic model accounts for, although indirectly, the strategic behavior of participating agents. By stochastically modeling the auction, the p-strategy agent derives the probability of success and the payoff of success, and trades off the probability against the payoff. These tradeoffs are manifested in its decision of raising (or dropping) and narrowing (or spreading) the offer prices. As knowing about each individual agent is a difficult task in the changing environment of the CDA, modeling the aggregate behavior of the agents (but not the behavior or internal reasoning of any individual agent) is a feasible solution for developing an agent bidding strategy. Using MCs to devise a strategy based on the aggregate population of buyers and sellers is novel, and demonstrates the value of the MC model to agent design. The MC model is reusable and thus the techniques of building the MC model for the auction can benefit other agent designers.

The p-strategy is a practical agent bidding strategy to be used on behalf of human bidders for the CDA. The complexity of the p-strategy is proportional to the number of MC states, which is bounded by the maximum number of standing offers. As a result, compared to modeling the internal reasoning of each individual agent (which requires exponential computational power as the number of agents increases), the complexity of the p-strategy does not increase with the number of agents in the auction.

We have empirically evaluated the performance of the p-strategy. First, we have compared the p-strategy seller to agents using other bidding strategies under various environments. The results indicate that the p-strategy outperforms other agent strategies in the CDA in a majority of experiments. In particular, the p-strategy seller performs well when many buy offers are available, as it can extract more profit per match at the expense of buyers. However, the performance of the p-strategy seller degrades with high competition among sellers for much smaller buy offers, or with multiple competing p-sellers. Second, we have analyzed the behavior of the p-strategy. When the negotiation zone is wider, the p-strategy seller raises its offer price to take advantage of the wide spread between the buy and sell offer prices. It also raises its offer price when there are many buy offers.

The experimental results and analyses contribute not only to our understanding of the performance of the p-strategy and the other strategies being compared but also to understanding of the behavior of an ideal agent bidding strategy. The good performance of the p-strategy seller is the result of its dynamic bidding behavior of raising or dropping the offer prices, which is possible by modeling the auction at runtime. The agent strategies based on simple rules of thumb (e.g., FM and CP strategies) are effective when there is not much room to gain profit to begin with. When the negotiation zone is wider or there are many buy offers, for example, they perform poorly, by failing to take advantage of potential profit gain. This emphasizes the benefit of having a model of the runtime states of the auction, which is particularly important for the CDA where the runtime auction status is continuously changing. The offer-price distributions confirm our claim of the importance of auction modeling by an agent strategy. The offer-price distribution of the p-seller is similar to that of the OPT-seller, and that is why the p-seller performs most closely to the OPT-seller. The evaluation is done primarily on the seller, but we conjecture that the findings can be applicable to the buyer as well, as the CDA domain exhibits symmetry between the buyer and the seller.

Our experience in designing and evaluating the p-strategy will be useful to other researchers who study agent strategies. Researchers can design an agent strategy based on the findings reported in this paper, and also use the p-strategy as one of the benchmark strategies to compare against the performance of their agent strategy. Our analysis shows that there is about a 20% gap between the performance of the p-strategy and that of the OPT-strategy. Although the ideal performance of the OPT-strategy may be hard to achieve, agent designers can either improve upon the current design of the p-strategy or develop new strategies to shrink the gap. More generally, the research reported in this paper contributes to the computational decision-making literature, and particularly the multi-agent reasoning literature, by giving a concrete example of where modeling the trajectory of the aggregate effects of agents' decisions (in this case, modeling the CDA) avoids the need to model the individual decision-making behaviors of the agents, and thus holds promise to simplify scaling up to larger agent populations.

From the experiments, we have learned that the p-strategy performs well in a majority of environments, particularly when other agents use naïve strategies, but its superiority diminishes as competition among sellers increases. Such observations prompt us to incorporate adaptation capability into the p-strategy. The idea is to let the agent figure out when using stochastic modeling is beneficial and when using other simpler strategies is beneficial at run time. Such a hybrid strategy, which we call the *adaptive p-strategy*, is our current ongoing work. The adaptive p-strategy agent will be able to capitalize on the strength of stochastic modeling (of the dynamics and uncertainties of the auction process) and avoid the shortcomings of stochastic modeling at the same time, by adaptively deciding when to use the p-strategy and when not to.

### Acknowledgements

We would like to thank the anonymous reviewers for their constructive comments. The organization, presentation, and many details of the paper reflect their suggestions. We also would like to thank Michael P. Wellman and Gary Olson for their feedback. This work has been funded in part by the joint NSF/DARPA/NASA Digital Libraries Initiative under CERA IRI-9411287, and by NSF grants IIS-9872057 and IIS-0112669.

### Appendix A: Notations

| $\rho$ | Offer price (sometimes $\rho_s$ and $\rho_b$ for the seller's and the buyer's offer prices, respectively) |
|---|---|
| $u(\rho)$ | Expected utility of offer $\rho$ |
| $P_S(\rho)$ | Probability of Success, given $\rho$ |
| $P_F(\rho)$ | Probability of Failure, given $\rho$ |
| $Payoff_S(\rho)$ | Payoff of Success, given $\rho$ |
| $Payoff_F(\rho)$ | Payoff of Failure, given $\rho$ |
| $C$ | Cost |
| $V$ | Valuation |
| $CP$ | Clearing price |
| $\Delta_S(\rho)$ | Delay between the offer and the successful match |
| $\Delta_F(\rho)$ | Delay between the offer and failure |
| $TD(\Delta(\rho))$ | Time discount for delay of $\Delta(\rho)$ $(-c.\Delta(\rho)$, when c < 0) |

**Table A.1: Notations used in defining the agent's decision problem in Section 3.**

| $b$ | The buyer's offer |
|---|---|
| $s$ | The seller's offer |
| $S^p$ | The p-strategy agent's offer |
| $(b b .. b s ... s)$ | Standing offers in the auction ordered from the lowest offer to the highest |
| $f_b(b)$ | Probabililty distribution of buyer's offer price |
| $f_s(s)$ | Probability distribution of seller's offer price |
| $cp$ | Clearing-price quote (compared to the actual clearing price (CP)) |
| $\$(i)$ | Agent $i$'s offer price (Note that $\$(s^p) = \rho$) |

**Table A.2: Notations used in building the MC model for CDA in Section 4.**

| $\boldsymbol{P}$ | Transition probability matrix ($\lVert P_{ij} \rVert$) |
|---|---|
| $P_{ij}$ | Transition probability from state $i$ to state $j$ |
| $P_{ij}^{(S)}$ | Conditional transition probability when the process ends up in $S$ |
| $P_{ij}^{(F)}$ | Conditional transition probability when the process ends up in $F$ |
| $\boldsymbol{M}$ | Fundamental matrix ($\lVert \mu_{ij} \rVert$) |
| $\mu_{ij}$ | Average number of visits the process makes to transient state $j$ starting from state $i$ before it enters one of the absorbing states |
| $\mu_{ij}^{(S)}$ | Average number of visits the process makes to transient state $j$ starting from state $i$ before it enters $S$ |
| $\mu_{ij}^{(F)}$ | Average number of visits the process makes to transient state $j$ starting from state $i$ before it enters $F$ |
| $f_{ij}$ | Probability the process enters absorbing state $j$ starting from state $i$ |
| $\omega_{ij}$ | Reward associated with the state transition from $i$ to $j$ (= the cost of delay) |
| $T$ | Set of transient states |
| $T^C$ | Set of absorbing states |

**Table A.3: Notations used in computing the utility value in Section 4.**